\documentclass{article}

\usepackage[preprint]{neurips_2025}

\usepackage[utf8]{inputenc} 
\usepackage[T1]{fontenc}    
\usepackage{hyperref}       
\usepackage{url}            
\usepackage{booktabs}       
\usepackage{amsfonts}       
\usepackage{nicefrac}       
\usepackage{microtype}      
\usepackage{xcolor}         
\usepackage{graphicx}  
\usepackage{fontawesome}
\usepackage{amsmath}
\usepackage{wrapfig}

\usepackage{float} 
\usepackage{subcaption} 

\definecolor{goldenorange}{RGB}{246,185,47}
\definecolor{customgray}{RGB}{156, 156, 156}

\newcommand{\frozenweights}[1]{\textcolor{customgray}{#1}}
\newcommand{\trainedweights}[1]{\textcolor{goldenorange}{#1}}

\title{Imagine for Me: Creative Conceptual Blending of Real Images and Text via Blended Attention}

\author{%
  Wonwoong Cho\thanks{Elmore Family School of Electrical and Computer Engineering, Purdue University}\\
  \And
  Yanxia Zhang\thanks{Toyota Research Institute} \\
  \And
  Yan-Ying Chen$^{\dagger}$ \\
  \And
  David I. Inouye$^{*}$ \\
}

\begin{document}

\maketitle

\begin{figure}[ht]
    \centering
    \vspace{-0.1in}
    \includegraphics[width=1\textwidth]{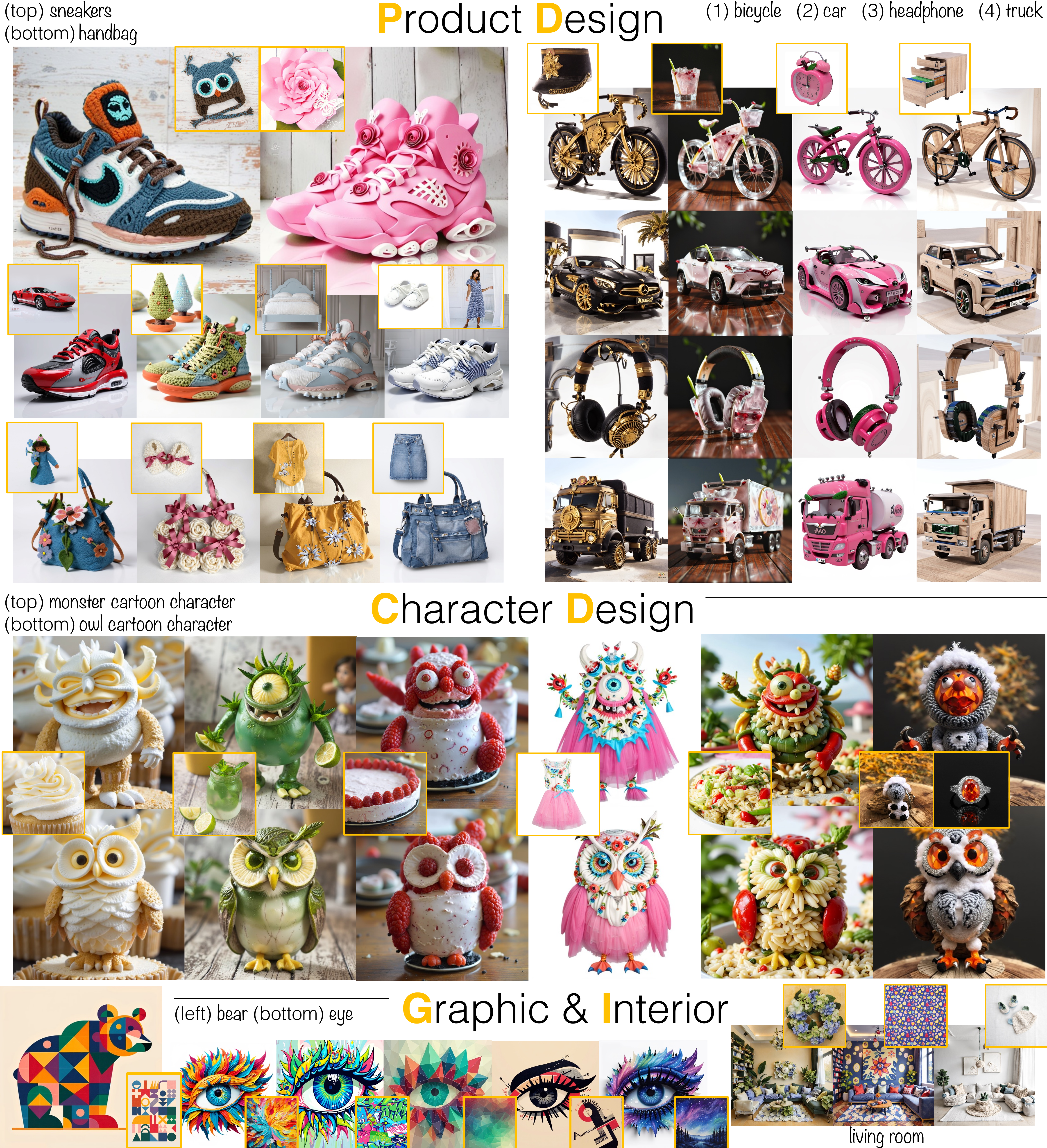}  
    \caption{Visual and textual conceptual blending results of IT-Blender based on FLUX.1-dev.}
    \label{fig:main 1page figure}
    \vspace{0.1in}
\end{figure}

\begin{abstract}
  Blending visual and textual concepts into a new visual concept is a unique and powerful trait of human beings that can fuel creativity. 
  However, in practice, cross-modal conceptual blending for humans is prone to cognitive biases, like design fixation, which leads to local minima in the design space.
  In this paper, we propose a T2I diffusion adapter ``IT-Blender'' that can automate the blending process to enhance human creativity.
  Prior works related to cross-modal conceptual blending are limited in encoding a real image without loss of details or in disentangling the image and text inputs. 
  To address these gaps, IT-Blender leverages pretrained diffusion models (SD and FLUX) to blend the latent representations of a clean reference image with those of the noisy generated image.
  Combined with our novel blended attention, IT-Blender encodes the real reference image without loss of details and blends the visual concept with the object specified by the text in a disentangled way.
  Our experiment results show that IT-Blender outperforms the baselines by a large margin in blending visual and textual concepts, shedding light on the new application of image generative models to augment human creativity. Our project website is: \trainedweights{\url{https://imagineforme.github.io/}}. 
\end{abstract}

\section{Introduction}
\label{sec:intro}

\textit{``Conceptual integration is at the heart of imagination'' --- \citet{fauconnier2008way}}

Conceptual integration/blending~\citep{fauconnier1998conceptual,fauconnier2008way,coulson2001semantic} is a theory in Cognitive Science, which can describe the human's cognitive process combining a visual and textual concepts into a new idea. It is one of the most essential virtues in the creative industries (e.g., product design, character design, fashion design, interior design, graphic design, art, and advertisement) because conceptual blending can provide inspirational and creative design ideas by creating new combinations or reinventing existing ones~\citep{gabora2002cognitive}. 

Prior works \citet{yang2009observations,hyun2018balancing,cai2023designaid} have shown that exploring the design concepts and space as much as possible can produce better design results especially during the early phase of the design process (e.g., Conceptual Design~\citep{otto2003product} and the SCAMPER method~\citep{eberle1996scamper} in Concept Generation~\citep{ulrich2016product}). 

However, there can be two challenges to perform cross-modal visual and textual conceptual blending in practice. First, human's creativity easily got stuck in the suboptimal as shown in design fixation (i.e., a tendency of a designer to overly adhere to a limited set of solutions)~\citep{jansson1991design} and Einstellung effect (i.e., a cognitive bias from past experiences or familiar solutions to a problem, preventing them from exploring better alternatives)~\citep{luchins1942mechanization}.

Second, cross-modal conceptual blending itself is not a trivial task. It can be achieved by \textit{selective projection} process determining what and where to integrate the given multiple concepts~\citep{fauconnier1998conceptual}. It involves the laborious process for identifying features in each condition and comparing the semantic correspondence to find a way to meaningfully blend them together. 




Recent significant advances of text-to-image (T2I) diffusion models~\citep{rombach2022high,saharia2022photorealistic} and their applications for adding an image condition led us to the question, \textit{``Can pretrained diffusion models be used for cross-modal conceptual blending to augment creativity?''} 

If so, it can be very useful by 1) providing numerous conceptual blending results to explore broader design possibilities and 2) automating the conceptual blending process to minimize the time required to manually illustrate all design ideas. For example, suppose that we want to come up with a creative product design for sneakers. Instead of struggling with imagining what to combine with and how to apply the selective projection, we can simply give a prompt like ``a photo of sneakers, creative design.'' and give a set of reference images with a target concept and appearance, e.g., a sport car image for ``sleek'' or any knitted items for ``warm'' and ``cozy'' (e.g., Fig.~\ref{fig:main 1page figure}). We may also apply the same style to the multiple objects (e.g., bicycle and car) or add multiple visual concepts to the generated results. Even random reference images can be used to provide a serendipitous inspiration.

The question is how to perform selective projection in diffusion models, which must be done to achieve cross-modal conceptual blending. We think the key is the attention module~\citep{vaswani2017attention} (which is one of the most crucial components of modern diffusion models~\citep{rombach2022high,blackforestlabs2024flux}) because its mechanism, comparing similarity and selectively applying the value, is conceptually close to the selective projection. 

Earlier work, such as IP-Adapter~\citep{ye2023ip} and BLIP-Diffusion~\citep{li2023blip} proposed encoder-based methods to incorporate a reference image into text-guided generation with additional training. 
Although they show decent performance in blending visual and textual concepts with a fast inference time, their methods are limited in 1) disentangling textual and visual conditions and 2) preserving the detailed visual concept of the reference image due to the dependencies on the text cross-attention module and an external image encoder. 



Meanwhile, RIVAL~\citep{zhang2023real} and StyleAligned~\citep{hertz2024style} have shown the potential of the pretrained self-attention module of the T2I diffusion models in blending cross-modal concepts. 
Although they showed impressive performance in disentangling cross-modal concepts and applying detailed visual concepts from their own denoising chain to another, their performance is limited when a real reference image is conditioned due to the distribution shift of the inversion chain~\citep{zhang2023real}. They also have a slower inference time than the encoder-based methods. 

Filling the gap between both baseline approaches, we propose a novel image adapter ``Image-and-Text Concept Blender'' (IT-Blender) that can imagine for us by blending cross-modal concepts with fast inference time. IT-Blender learns to blend visual concepts from a real image without loss of details, in a disentangled manner from the textual concept (i.e., text determines semantics while a reference image determines visual concepts such as texture, material, color, and local shape).


Briefly, instead of using an external image encoder, we leverage the denoising network as an image encoder to maintain the details of visual concepts. 
As opposed to recent related literature without an external image encoder~\citep{wu2025less,tan2024ominicontrol}, our proposed method does not have any architectural dependency (i.e., applicable to both UNet-based~\citep{rombach2022high} and DiT-based diffusion models~\citep{blackforestlabs2024flux}).
We design a novel \textit{Blended Attention} on top of the self-attention module, where detailed visual concepts can be preserved, and textual concepts are physically separated, encouraging disentanglement of textual and visual concepts. Blended Attention is trained to be specialized in finding a semantic correspondence between two latents; one from the real reference image and the other from the generated image.


Our baseline experiment results on disentanglement, concept preservation, and blending score (in Appendices) demonstrate that IT-Blender outperforms the baselines in cross-modal conceptual blending in both UNet-based (SD 1.5) and DiT-based (FLUX) architectures.

\vspace{-0.01in}
\section{Related Works}
\label{sec:related works}
\vspace{-0.01in}
In this section, we introduce previous studies related to visual-and-textual conceptual blending, based on diffusion models~\citep{ho2020denoising,dhariwal2021diffusion,songscore}. 



\textbf{Applications for spatially aligned control.} Prior works~\citep{zhang2023adding,mou2024t2i,hertz2022prompt,tumanyan2023plug,liu2024towards} have achieved impressive performance in spatially aligned control. However, their methods are mainly designed for local photo editing instructed by text, which is not suitable for our conceptual blending task to augment creativity.


\textbf{Applications based on text cross-attention module related to conceptual blending.} IP-Adapter~\citep{ye2023ip}, BLIP-Diffusion~\citep{li2023blip}, and ELITE~\citep{wei2023elite} are closely related to conceptual blending task. They proposed an adapter based on text cross-attention module to encode and incorporate reference image information into the text-guided image generation process. Even though decently working for cross-modal conceptual blending, their methods are limited in two aspects. First, the encoder-based methods often fail in disentangling visual and textual concepts. This is because a reference image relies on the text cross-attention module, potentially entangling the cross-modal information. Second, encoder-based methods are limited in blending the detailed visual concepts because of a dependency on an external image encoder, where visual details can be lost.


\textbf{Applications of self-attention module related to conceptual blending.}
Self-attention module is shown to be effective in combining two spatial features. RIVAL~\citep{zhang2023real} and StyleAligned~\citep{hertz2024style} proposed to modify the self-attention module to be a sort of cross-attention form. Starting from the noise corresponding to the real reference image through inversion methods~\citep{song2020denoising,mokady2023null}, they combine a denoising chain with an inversion chain to blend the spatial features. Although they can blend cross-modal concepts without training, their methods are inherently limited when a real reference image is given, due to the distributional gap between the latents from the inversion chain and the denoising chain~\citep{zhang2023real}.
Similar ideas are used in \citep{cao2023masactrl,alaluf2024cross}, but their methods are specifically designed for non-rigid image editing or a blending of two visual concepts in a disentangled manner.

\textbf{Transformer-based applications related to conceptual blending.}
UNO~\citep{wu2025less}, OminiControl~\citep{tan2024ominicontrol}, and IC-LoRA~\citep{huang2024context} have shown impressive performance in subject-driven image generation by leveraging diffusion transformers as image encoder. However, their sequentially concatenating methods are only applicable to Diffusion Transformers. 
Moreover, their methods are not suitable for conceptual blending because of the strong subject preservation.

\noindent \textbf{Downstream tasks of conceptual blending.} 
The notion of conceptual blending to generate an image has been widely explored in various tasks such as image stylization and creative object generation.

For image stylization, StyleDrop~\citep{sohn2023styledrop} and other related works~\citep{voynov2023p+,zhang2023inversion,shah2024ziplora,frenkel2024implicit} propose to blend textual and visual concepts in a disentangled manner by personalizing the input tokens or finetuning the networks (e.g., additional adapter or LoRA~\citep{hu2022lora}). However, they are limited in scalability, as optimization is required for each visual concept.

For creative object generation, most prior work~\citep{richardson2024conceptlab,li2024tp2o,feng2025redefining} has primarily addressed unimodal blending of textual concepts. MagixMix~\citep{liew2022magicmix} and ATIH~\citep{xiong2024novel} are designed for cross-modal conceptual blending, but their methods are designed for spatially aligned control, which is different from our task. ATIH also requires iterative optimization steps for each reference image, which limits its scalability in practice.

More importantly, we argue that conceptual blending is a more fundamental and broad notion and is not limited to a specific task. As shown in the results in Section~\ref{appendix subsec: additional results}, once trained, IT-Blender is applicable not only to image stylization and creative object generation, but also to a variety of design tasks without additional per-instance optimization.

\noindent \textbf{Generative models augmenting human creativity.} 
Previous studies~\citep{franceschelli2024creativity,hwang2022too} have shown the potential of generative models for augmenting creativity. \citet{cai2023designaid} proposed a diffusion framework to diversify image generations to provide inspiration for designers.  CreativeConnect~\citep{choi2024creativeconnect} proposed generative AI pipelines that can help graphic designers to have more design ideas by reference recombination process.
Creative Blends~\citep{CreativeBlends25} proposed a system that takes multiple textual concepts as input from users and outputs an image with the blended concepts. The conducted user study shows that visualizing these blended concepts can reduce cognitive load for participants and also foster creativity.

\vspace{-0.05in}
\section{Method}
\vspace{-0.05in}
\label{sec:method}

In this Section, we describe our proposed method (IT-Blender) that adapts the pretrained projection layers of self-attention module to the visual and textual conceptual blending task.
In Section~\ref{subsec: preliminaries},
we first describe the preliminaries of the T2I diffusion models. 
In Section~\ref{subsec:Image and Text Blender (IT-Blender)}, we introduce IT-Blender with a novel blended attention module that can blend the visual concept of a reference image into the text-guided generation process with enhanced semantic correspondence retrieval for the real image. 
 
\subsection{Preliminaries}
\label{subsec: preliminaries}
\textbf{StableDiffusion and FLUX.}
StableDiffuison (SD)~\citep{rombach2022high} is widely used open source diffusion models for T2I synthesis. SD is trained with a denoising objective~\citep{ho2020denoising}, and UNet~\citep{ronneberger2015u} is used as denoising networks.
FLUX~\citep{blackforestlabs2024flux} is advanced diffusion models based on diffusion transformers (DiT)~\citep{peebles2023scalable}, which is trained with a score matching objective. 
SD 1.5 and FLUX.1-dev are used in our experiment.

\textbf{Self-Attention module and its application.} Self-attention (SA) module~\citep{zhang2019self,rombach2022high,blackforestlabs2024flux} is one of the most important components of modern diffusion models. It not only learns to capture long-range dependencies, but also learns to encode spatial representations optimized for similarity comparison; what to aggregate and what to ignore based on semantic correspondence of the input itself. In our paper, the projection layers $\frozenweights{W_Q}$, $\frozenweights{W_K}$, and $\frozenweights{W_V}$ are pretrained weights of SA module. For brevity, we omit the layer notation for the projection layers.


As mentioned earlier, \citet{zhang2023real,hertz2024style} have shown that
visual concept of a reference image can be blended in the generation process of pretrained T2I models without additional training.
The detailed methodologies differ, but conceptually they suggested image Cross Attention (imCA) between two latents; $Z_{\text{noisy}}$ from a denoising chain and $Z_{\text{inv}}$ from an inversion chain, i.e., 
$\text{imCA}(Z_{\text{noisy}},Z_{\text{inv}})=\text{imCA}(Z_{\text{noisy}},Z_{\text{inv}};\frozenweights{W_Q},\frozenweights{W_K},\frozenweights{W_V}).$ This indicates the key and value of the SA module from $Z_{\text{inv}}$ are combined with the query of the SA module from $Z_{\text{noisy}}$, i.e., $\sigma\left( (Z_\text{noisy}\frozenweights{W_Q})(Z_\text{inv}\frozenweights{W_K})^T/\sqrt{d_k} \right) Z_\text{inv}\frozenweights{W_V}$. Note that the operation of imCA is essentially a cross-attention mechanism, but used in a distinct way, i.e., cross-attention in SA layers with $\frozenweights{W_Q}$,$\frozenweights{W_K}$,$\frozenweights{W_V}$.

\subsection{Image and Text Blender (IT-Blender)}
\label{subsec:Image and Text Blender (IT-Blender)}

\textbf{Setup and overview.}
We aim to generate an image where cross-modal concepts from a given real image and a text prompt are naturally blended without loss of details, in a disentangled manner. As mentioned in Section~\ref{sec:intro}, attention module can be a key to implement the conceptual blending process. 

One key observation for the inversion-based imCA approaches is that: they have the advantage in applying the details of the visual concepts in a disentangled manner, while the performance is degraded when real images are given as input due to the distribution shift of the inversion chain. Hence, our goal is to have a real image adapter that is trained to incorporate a given reference image into the pretrained projection space of the SA module. 
Since textual concepts are constantly provided through the text CA modules (which are physically separated from the SA modules), IT-Blender aims to blend visual concepts from the reference image with the text-guided generation process.


\begin{wrapfigure}{l}{0.5\textwidth}
  \centering
  \includegraphics[width=0.5\textwidth]{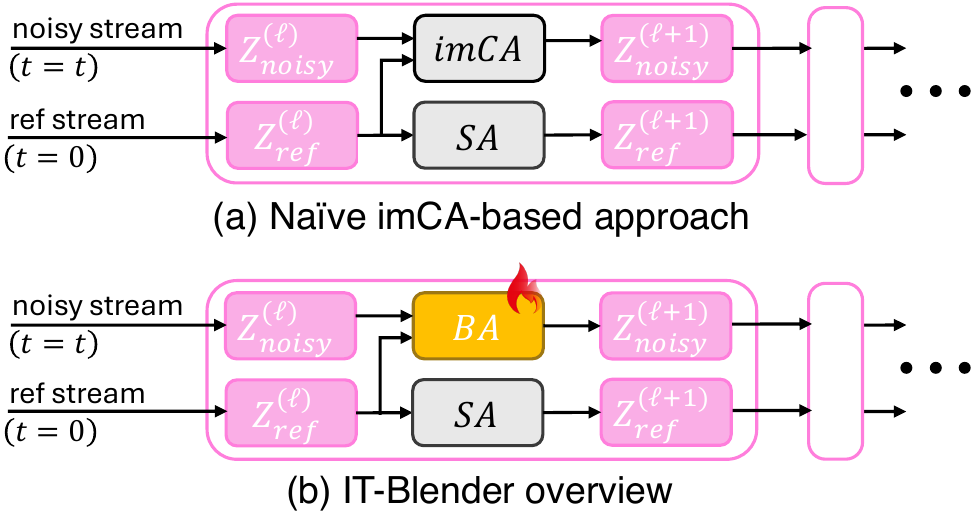}
  \caption{IT-Blender overview}
  \label{fig:it_blender_overview and imCA}
  \vspace{-0.25in}
  
\end{wrapfigure}

Our method only trains the newly introduced adapter parameters while freezing all the pretrained weights, similar to prior works~\citep{mou2024t2i,zhang2023adding,ye2023ip,tan2024ominicontrol,wu2025less}. The denoising objective is used for SD1.5~\citep{rombach2022high} and the denoising score matching objective is used for FLUX~\citep{blackforestlabs2024flux}. 


The challenges are 1) how to encode a real image without loss of details, and 2) how to blend the encoded real image feature into the projection space of the pretrained SA module.


\textbf{Native image encoding.}
Interestingly, diffusion models already know how to encode a real image $X_{\text{ref}}$ into the denoising networks. It can be simply achieved by forwarding a clean version of $X_{\text{ref}}$ with $t=0$. This provides a sequence of latent representations across the $L$ layers of the denoising networks: $(Z_{\text{ref}}^{(1)},Z_{\text{ref}}^{(2)},...,Z_{\text{ref}}^{(L)})$. This representation has some similarities to the inversion methods in which there is a latent representation at every layer of the network for each denoising step. However, it is fundamentally different because it's timestep is set to 0 for all denoising steps, i.e., the clean latent representations can be used at every timestep. We hypothesize that these clean representations are more helpful for conceptual blending because they encode the details of the clean image rather than noisy images as in inversion-based methods. Furthermore, our approach does not require image inversion, which is computationally expensive. 

Despite the benefits of this clean representation, it is unclear how to incorporate a set of clean latent features per layer from the denoising networks into the regular denoising process.
One simple naïve approach inspired by prior works is to simply use an imCA module to blend the clean reference latent $Z_{\text{ref}}$ into the noisy latent $Z_{\text{noisy}}$, i.e., replace $\text{SA}(Z_{\text{noisy}})$ modules with $\text{imCA}(Z_{\text{noisy}},Z_{\text{ref}})$, as shown in Fig.~\ref{fig:it_blender_overview and imCA} (a). While in theory this could be done without retraining by using the pretrained self-attention module weights as in 
\citet{hertz2024style,zhang2023real}, the performance would be poor because of a significant distribution shift; the reference latents are from a clean image with $t=0$ while the noisy latents are from noisy images with a $t \geq 0$. Fig.~\ref{fig:mask results and uno} (a) shows the empirical verification of the hypothesis. 
Thus, a new blending module and finetuning method is needed that can use the clean latents but seamlessly blend the visual concept information into the noisy latents.
 

\textbf{IT-Blender.} To bridge the gap, we design IT-Blender to have our novel blended attention (BA) module with trainable parameters that can learn how to map the clean $Z_{\text{ref}}$ to the $Z_{\text{noisy}}$ in the projection space. 

 \begin{wrapfigure}{r}{0.4\textwidth}
  \centering
  
  \vspace{-0.1in}\includegraphics[width=0.4\textwidth]{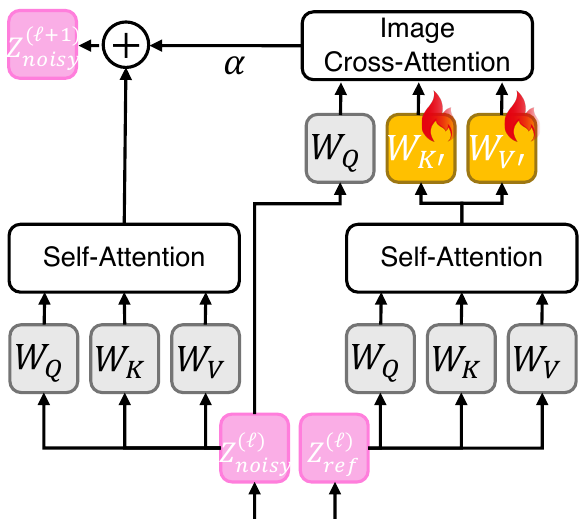}
  \caption{Blended attention at $\ell$-th layer.}
  \label{fig:blended attention (BA)}
  \vspace{-0.25in}
\end{wrapfigure}

As shown in Fig.~\ref{fig:it_blender_overview and imCA} (b), IT-Blender has two streams; noisy stream and reference stream. The noisy stream refers to the regular denoising chain from $t=T$ to $t=0$ during sampling or randomly sampled $t$ during training. The reference stream is for encoding a reference image without any noise. Along this stream, $t=0$ is constantly given for both training and sampling. The same text prompt is used for both streams. The training objective is applied only to the noisy stream.

\textbf{Blended Attention (BA).} 

As shown in Fig.~\ref{fig:blended attention (BA)}, we design blended attention to have a residual structure with two terms; the first term on the left is the original pretrained self-attention module, which can keep the estimation on the original trajectory. The second imCA term on the right is the key to blended attention, which enables a blending of visual and textual concepts by bridging the clean reference stream with the noisy stream in the output space of the SA module. The $\ell$-th self-attention layers of the denoising networks are changed to our blended attention as shown in the equation below:
\begin{equation}
    \text{BA}=\text{SA}(Z_{\text{noisy}}^{(\ell)}) + \alpha \, \text{imCA}(Z_{\text{noisy}}^{(\ell)},\text{SA}(Z_{\text{ref}}^{(\ell)});\frozenweights{W_Q},\trainedweights{W_{K'}},\trainedweights{W_{V'}}),
    \label{eq:blended attention}
\end{equation}
where $\trainedweights{W_{K'}}$ and $\trainedweights{W_{V'}}$ are trainable parameters. The layer notation for the projection layers is omitted for the brevity purpose. $\alpha$ is set to be 1 during the training while set to be a constant $< 1$ during sampling. In our experiments, we empirically used $\alpha=0.25$ for SD and $\alpha=0.6$ for FLUX (the visualization of varying $\alpha$ s is shown in Fig.~\ref{appendix fig:varying_alpha}). For training, $\trainedweights{W_{K'}}$ and $\trainedweights{W_{V'}}$ are randomly initialized.

The imCA term in Eq.~\ref{eq:blended attention} plays a role in dynamically aligning $\text{SA}(Z_{\text{ref}}^{(\ell)})$ to $\text{SA}(Z_{\text{noisy}}^{(\ell)})$ in the output space of SA by optimizing $\trainedweights{W_{K'}}$ and $\trainedweights{W_{V'}}$ to fetch the useful visual information to denoise from the reference stream, driven by the query from the noisy stream.

\vspace{-0.1in}
\section{Experiments}
\label{sec:experiments}
Detailed experiment settings and implementation details are provided in Section~\ref{appendix sec:experiment settings} and~\ref{appendix sec:implementation details}.

\textbf{Data.} For training and testing SD 1.5 and FLUX, we used a squared subset of LAION2B-en-aesthetic dataset~\citep{opendiffusionai_laion2b_en_aesthetic_square,schuhmann2022laion}, which contains around 300k squared images (with at least a resolution of 1,024$\times$ 1,024) and their paired text prompt.

\textbf{Metrics for baseline comparison.} We mainly evaluate how well the textual and visual concepts are disentangled. In our cross-modal blending task, semantics (i.e., object) must be determined by a text prompt, and visual concepts (e.g., texture, ingredient, material, color, and local shapes) need to be determined by a reference image. If visual and textual concepts are disentangled well, each of them should maintain high consistency after being blended with different combinations.
Therefore, the key to the evaluation is to measure set consistencies for visual concept and the textual concept, respectively. 
To measure the textual set consistency, we compare a set of generated samples with a fixed text prompt but with different reference images. The generated object must be consistent, and thus we used CLIP~\citep{radford2021learning} to measure the semantic similarity between all pairs of the generated images with a fixed prompt. 
To measure the visual set consistency, we compare the generated samples with a fixed visual prompt but with different text prompts. DINO is used to focus more on pure visual similarity, not semantics, following previous studies~\citep{ruiz2023dreambooth,hertz2024style}. Next, we measure the correct class predictions to measure whether the generated results preserve the textual concept. ChatGPT4.1~\citep{openai2023gpt4} is used.
For SD evaluation, 200 unseen samples with 30 text prompts are used (6,000 samples per baseline in total). For FLUX evaluation, 200 unseen samples with 20 text prompts are used (4,000 samples per baseline in total).
We also report the blending score and analysis in Section~\ref{appendix subsec: blending score by chatgpt}.  

\subsection{Baseline Comparison (SD)}

\textbf{Baselines.} To compare the performance of cross-modal conceptual blending in SD, we use two encoder-based methods (BLIP-Diffusion~\citep{li2023blip} and IP-Adapter~\citep{ye2023ip}) and two inversion-based methods (RIVAL~\citep{zhang2023real} and StyleAligned~\citep{hertz2024style}). We used SD 1.5 for all the baselines while SDXL~\citep{podell2023sdxl} is used in StyleAligned~\citep{hertz2024style} as their performance in SD 1.5 is worse by a large margin. 

 \begin{figure}[t]
 \vspace{-0.25in}
    \centering
    \includegraphics[width=\textwidth]{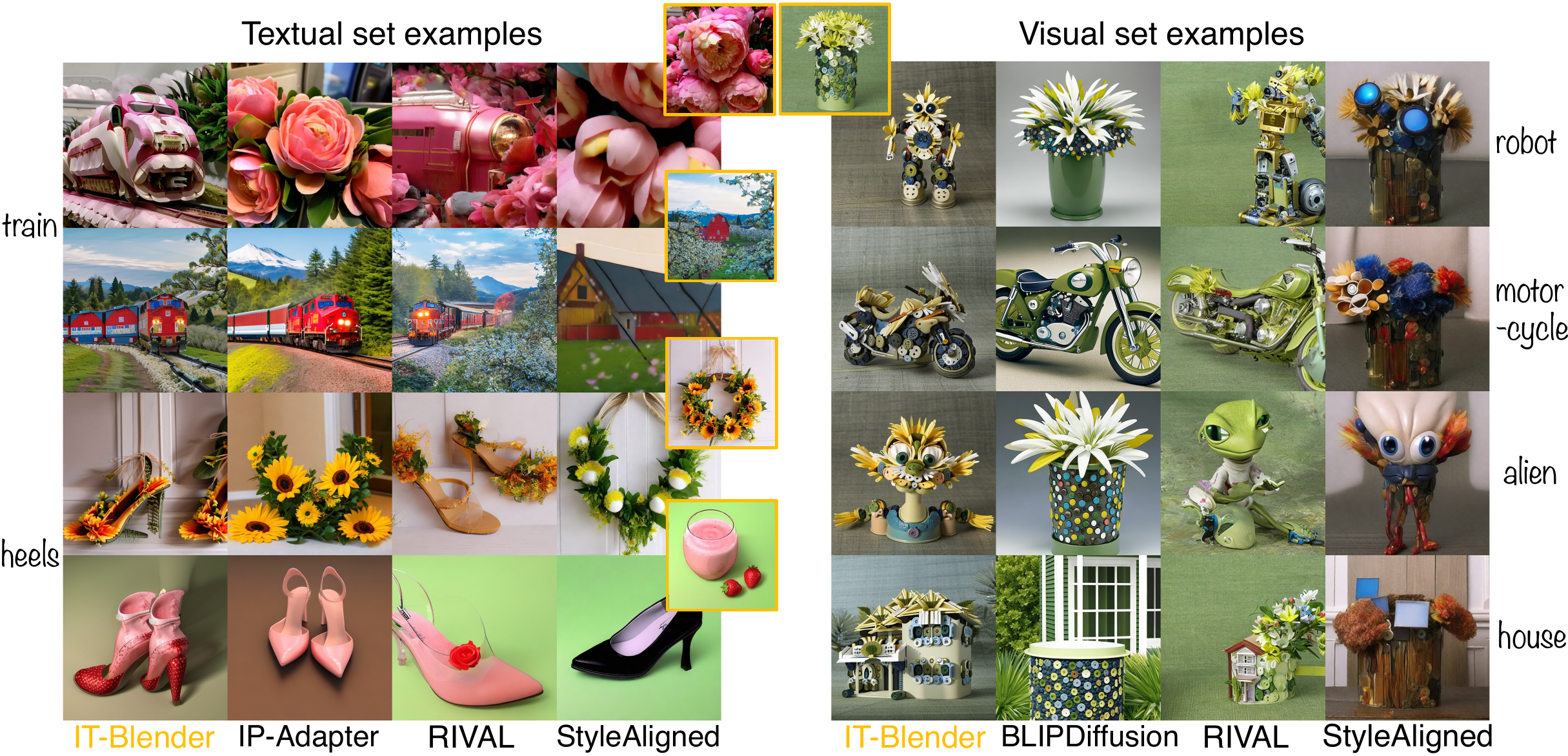}  
    \caption{Qualitative comparisons with the baselines in StableDiffusion. For each column of the textual set examples, every two row with the same text prompt need to be semantically consistent. Each column of the visual set examples need to be visually consistent.}
    \label{fig:sd15 qualitative visualization}
    \vspace{-0.15in}
\end{figure}

\begin{wrapfigure}{r}{0.35\textwidth}
  \begin{minipage}{\linewidth}
    \centering
    \vspace{-0.1in}
    \includegraphics[width=\linewidth]{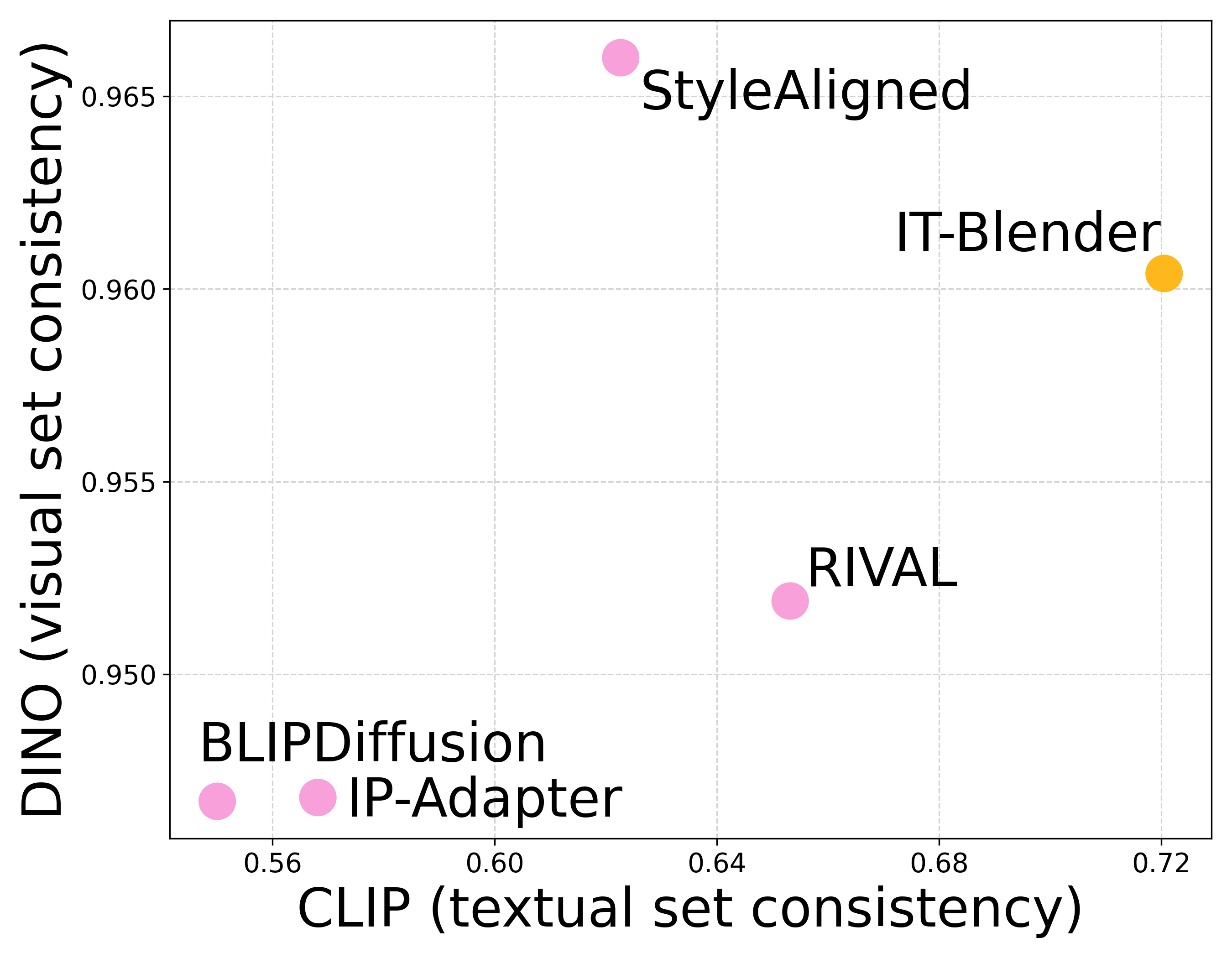}
    \vspace{5pt}
    \includegraphics[width=\linewidth]{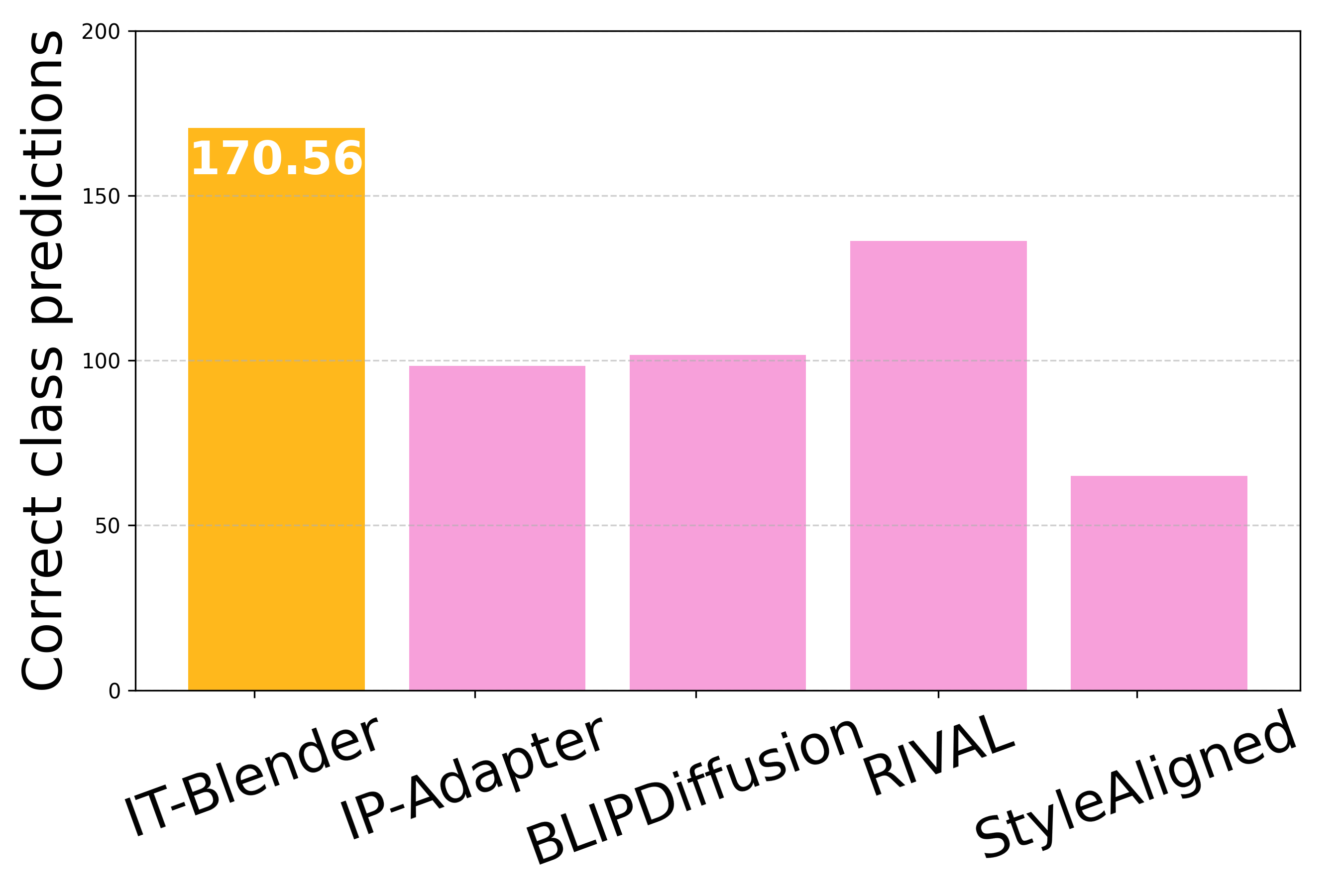}
  \end{minipage}
  \vspace{-0.15in}
  \caption{Visualizations of the quantitative comparison with the SD 1.5 baselines.}
    \label{fig:sd15 quantitative visualization}
    \vspace{-0.25in}
\end{wrapfigure}

\textbf{Results.} Both encoder-based baselines (IP-Adapter and BLIP-Diffusion) show similar patterns. First, the visual concept frequently dominates the generation process, and thus the generated images sometimes do not look like the object given as a text prompt (e.g., the flower train of IP-Adapter and the robot of BLIP-Diffusion in Fig.~\ref{fig:sd15 qualitative visualization}).
The same pattern is observed in quantitative evaluations. The encoder-based baselines show the lowest visual and textual set consistencies (Fig.~\ref{fig:sd15 quantitative visualization} top). This is because they frequently miss the textual concept, yielding inconsistency of the textual and visual sets. 

Similarly, the classification results in Fig.~\ref{fig:sd15 quantitative visualization} bottom show that IP-Adapter and BLIP-Diffusion often miss the target object; out of 200 samples, on average over the prompts, only around 100 samples are classified as a target object. We believe this is because their methods rely on the text CA module, which can inherently limit the disentanglement of visual and textual concepts.




Second, when the textual concept is properly applied, the generated results from IP-Adapter and BLIP-Diffusion often lose the details of the visual concept (e.g., the strawberry heels of IP-Adapter and the motorcycle of BLIP-Diffusion in Fig.~\ref{fig:sd15 qualitative visualization}). Additional DINO similarity experiments between a generated image and a reference image (IT-Blender (\textbf{0.837}), IP-Adapter (0.812), and BLIP-Diffusion (0.821)) support the observations. This is because IT-Blender does not rely on an external image encoder, while natively encodes images with the denoising networks, retaining visual details better.

As for inversion-based baselines, StyleAligned 
frequently misses the textual concept, as shown in the motorcycle and house examples in Fig.~\ref{fig:sd15 qualitative visualization}. The lowest classification score in Fig.~\ref{fig:sd15 quantitative visualization} bottom also quantiatively supports the observation. RIVAL shows worse performance than IT-Blender in both textual and visual set consistencies. This is because their inversion-based method is not specialized in retrieving semantic correspondence between the reference and the generated images, and thus the visual concepts are inconsistently applied to the generated images given varying inputs.


IT-Blender shows good performance in blending visual and textual concepts in a disentangled manner, as shown in the second-best visual set consistency and the best textual set consistency. The highest class prediction also supports the strong performance of IT-Blender in rigidly applying textual concepts. 
The superior disentanglement performance of IT-Blender is attributed to 1) self-attention-based design, which separates the visual and textual concepts, and 2) strong semantic correspondence retrieval by blended attention, with which the given visual concepts can be consistently applied given varying inputs.

Additional baseline comparisons are provided in Section~\ref{appendix sec: additional baseline comparisons}, e.g., ``blending score'' by ChatGPT and occasional unrealistic generations of inversion-based baselines in SD.



\subsection{Baseline Comparison (FLUX)}

\textbf{Baselines.} To compare the cross-modal conceptual blending performance in FLUX, we used three open-source baselines; IP-Adapter~\citep{ye2023ip}, OminiControl~\citep{tan2024ominicontrol}, and UNO~\citep{wu2025less}. For IP-Adapter, among two popular open source implementations, we used InstantX implementation~\citep{flux-ipa} as it is much better in blending visual and textual concepts. Both OminiControl and UNO are designed for subject-driven image generation by training additional lora modules on top of the pretrained FLUX. The experiment results of IP-Adapter and UNO are based on FLUX.1-dev while OminiControl is based on FLUX.1-schnell.

 \begin{figure}[t]
    \centering
    \vspace{-0.3in}
    \includegraphics[width=\textwidth]{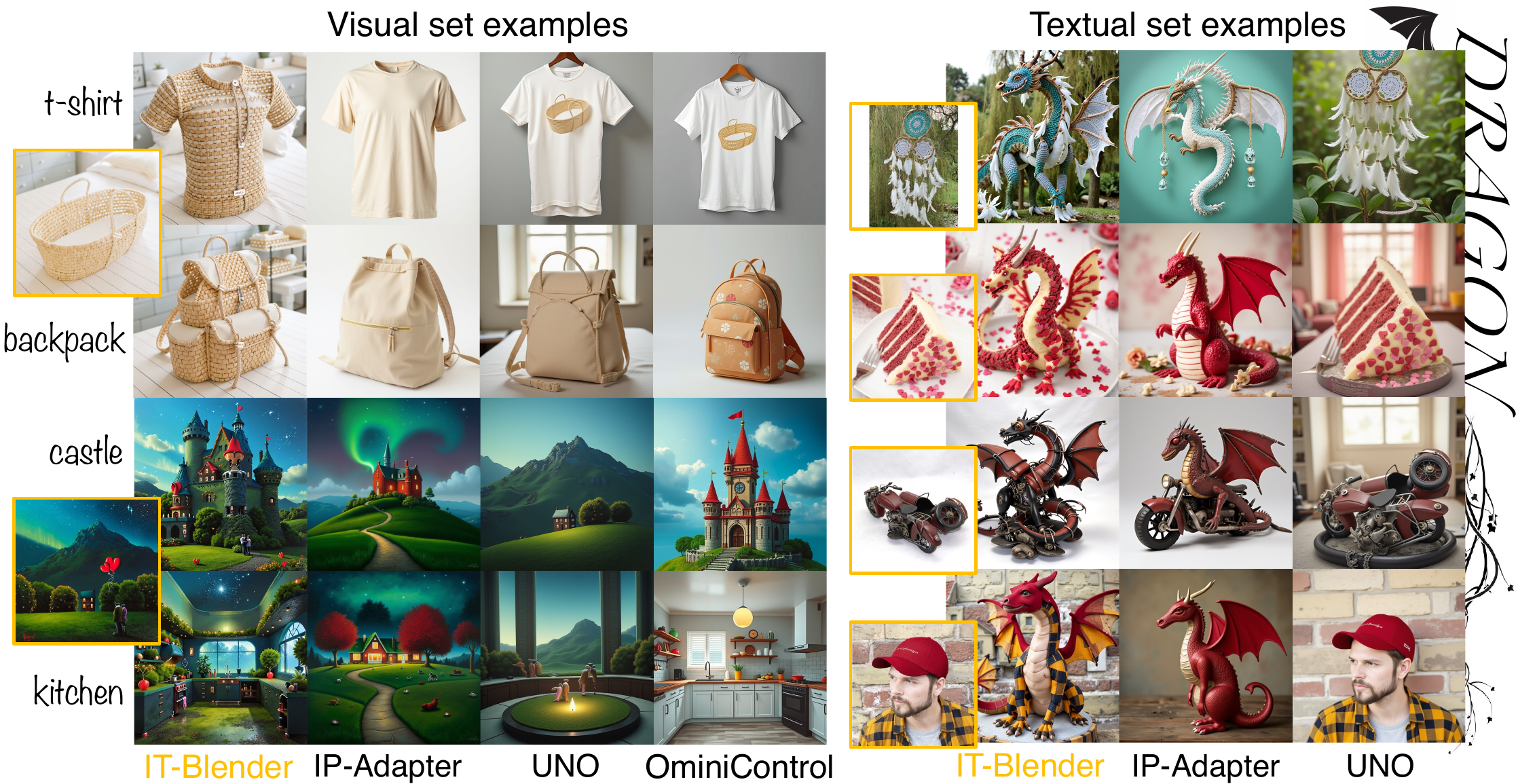}  
    \caption{Qualitative comparisons with the baselines in FLUX.}
    \vspace{-0.2in}
    \label{fig:flux qualitative visualization}
\end{figure}

\begin{wrapfigure}{r}{0.35\textwidth}
  \begin{minipage}{\linewidth}
    \centering
    \vspace{-0.2in}
\includegraphics[width=\linewidth]{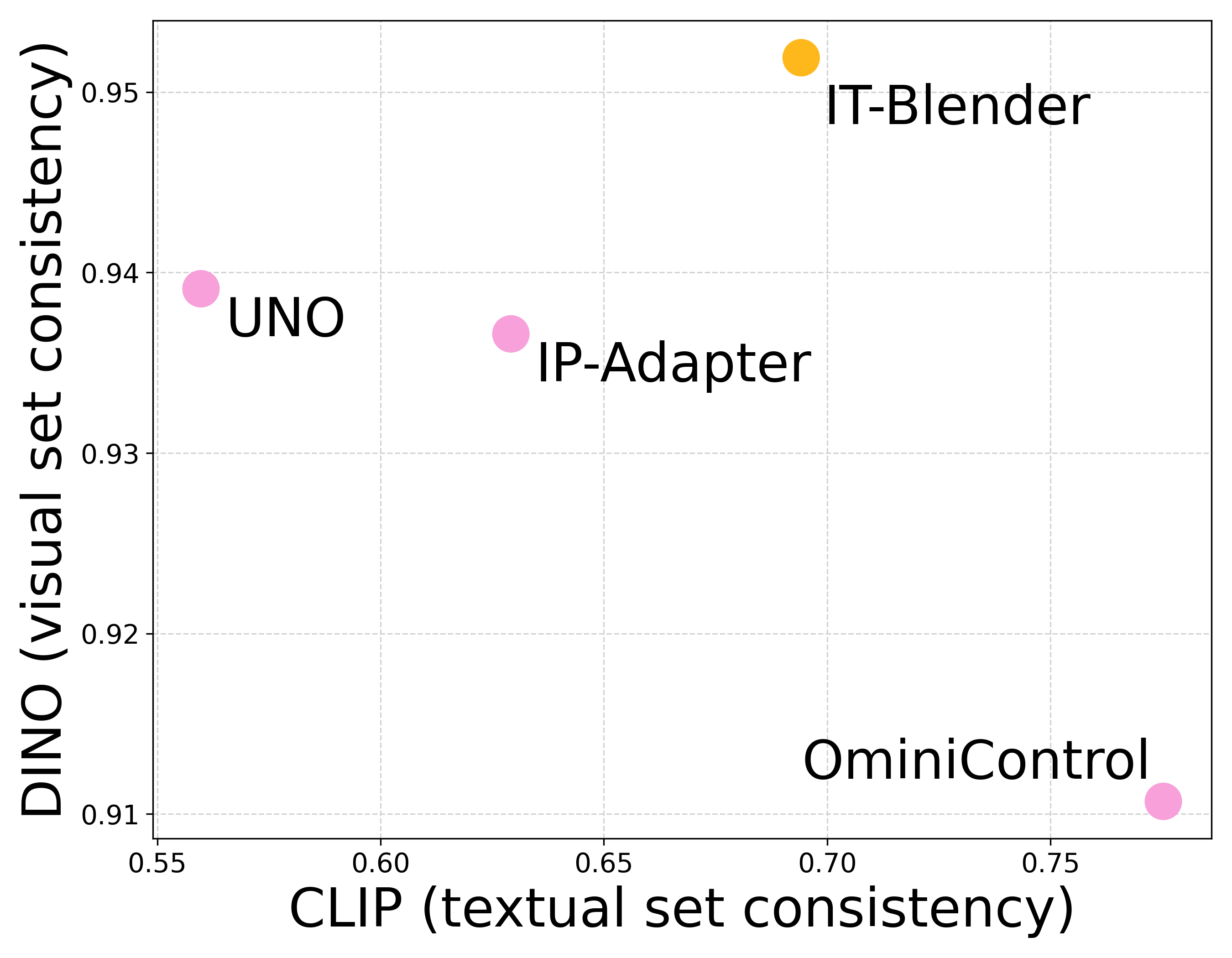}
    \vspace{5pt}
    \includegraphics[width=\linewidth]{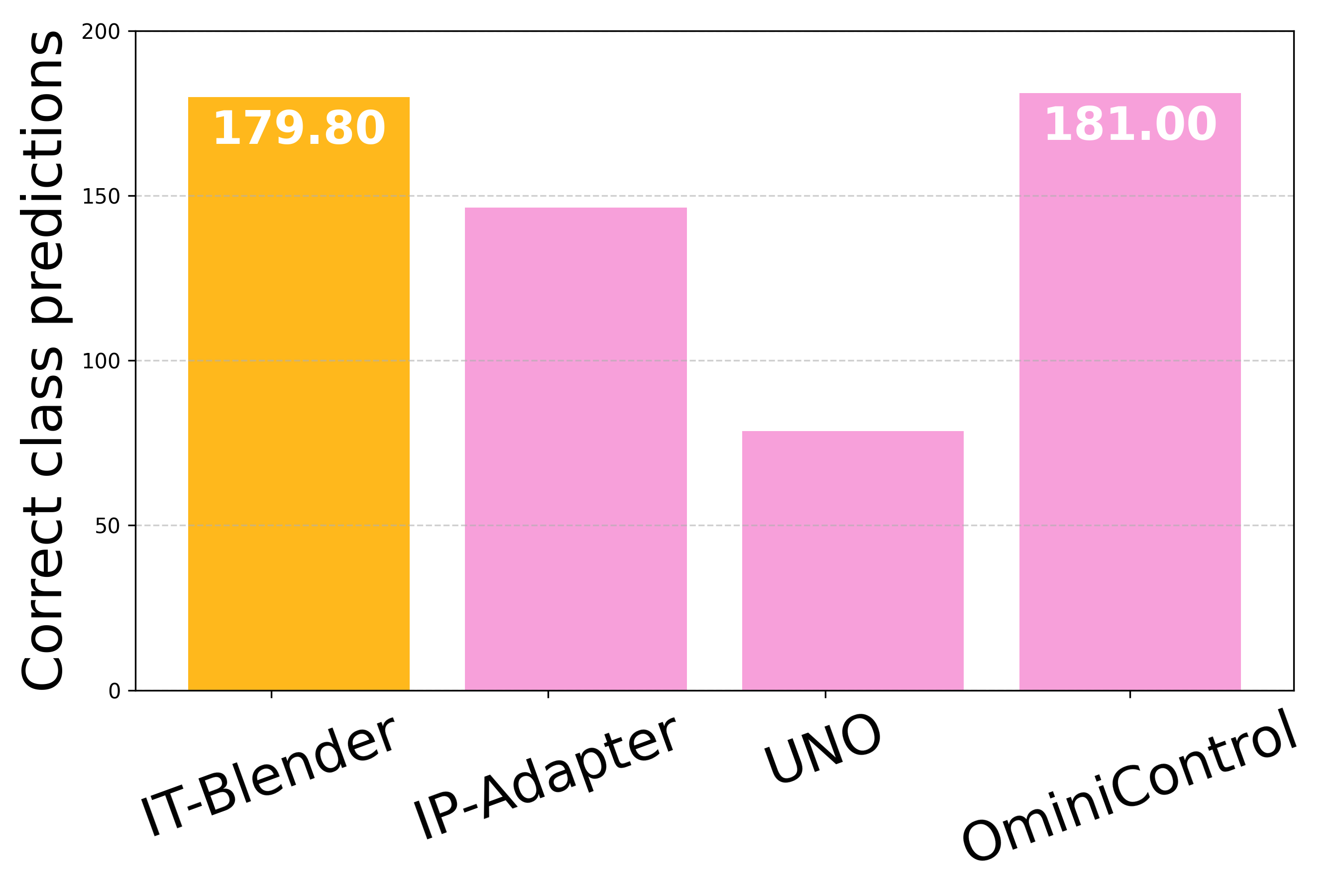}
  \end{minipage}
  \vspace{-0.15in}
\caption{Visualizations of the quantitative comparison with the FLUX baselines.}
    \label{fig:flux quantitative visualization}
    \vspace{-0.2in}
\end{wrapfigure}

\textbf{Results.} As UNO and OminiControl are specifically trained for subject-driven image generation with paired data, their models are not suitable for blending visual and textual concepts, especially when given visual and textual conditions are not highly correlated. As can be seen in Fig.~\ref{fig:flux qualitative visualization}, UNO and OminiControl show strong reference preservation, as shown in the basket-printed t-shirts. However, OminiControl often fails in incorporating the visual concept from the reference image (e.g., the backpack and kitchen examples), while UNO often fails in incorporating the textual concept (e.g., the castle and kitchen examples). IP-Adapter decently blends the visual and textual concepts, but they miss the details of the visual concepts (e.g., the dragons in the second and the fourth rows).

We also observe the similar patterns in the quantitative experiment results. OminiControl shows strong text guidance effect (e.g., the highest textual set consistency and classification in Fig.~\ref{fig:flux quantitative visualization}) while inconsistent reference image effect (e.g., the lowest visual set consistency). UNO shows relatively robust performance in preserving the given object in our task, as shown by the high visual set consistency score. However, the given text prompt is often ignored, which is shown by the lowest textual set consistency.
IP-Adapter demonstrates lower visual and textual set consistencies compared to ours, simialr to the SD experiment results. Compared to the baselines, IT-Blender shows the second-best textual set consistency and the best visual set consistency, showing superior performance in cross-modal conceptual blending.



\vspace{-0.05in}
\subsection{Ablation Study and New Applications}
\vspace{-0.05in}
In this section, we present interesting applications and visualize the attention mask to better understand what IT-Blender learns. More results are provided in the Appendices (e.g., applying multiple visual concepts in Section~\ref{appendix sec: multiple visual concepts} and more interesting results in Section~\ref{appendix sec: additional results}).

\begin{figure}[t]
 \vspace{-0.4in}
    \centering
    \includegraphics[width=\textwidth]{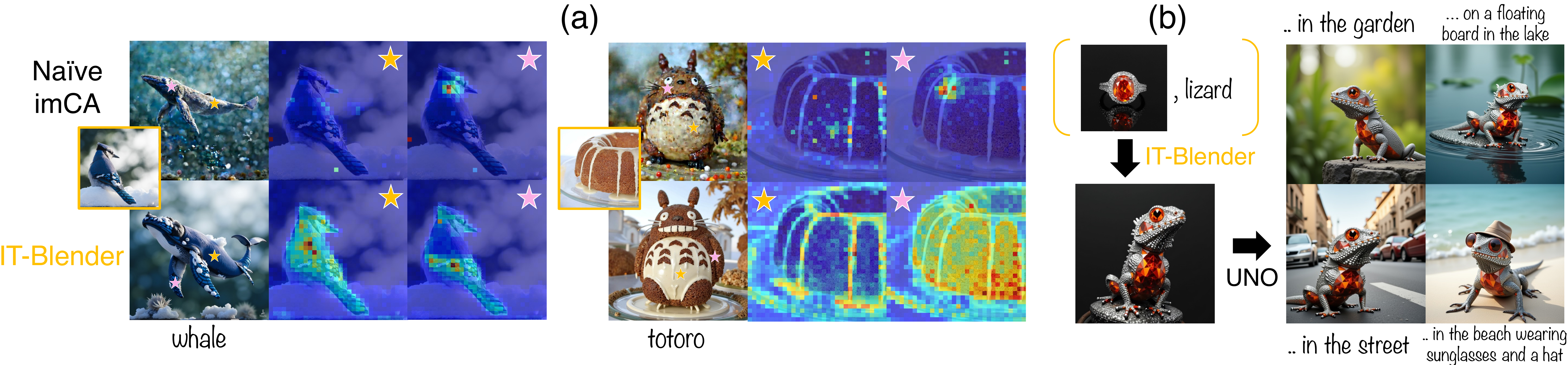}  
    \caption{(a) attention mask visualization of IT-Blender and naïve imCA (Fig.~\ref{fig:it_blender_overview and imCA} (a)). (b) our blended results can be applied to subject-driven generative models to create interesting novel scenes.}
  \label{fig:mask results and uno}
  \vspace{-0.1in}
\end{figure}

\textbf{Effects of the blended attention module.}  To intuitively understand what blended attention learns, we visualize the self-attention mask of BA modules in FLUX. Fig.~\ref{fig:mask results and uno} (a) shows the results. The attention masks of IT-Blender captures the visually corresponding texture area from the reference image. For example, the yellow star from the whale example mostly captures the fur area of the bird in the reference image, while the pink star mostly captures the feather area. However, the attention mask of the naive imCA-based approach (Fig.~\ref{fig:it_blender_overview and imCA} (a)), does not capture a meaningful area, and thus the generated results are also significantly degraded. This verifies our hypothesis that the distribution shift between clean $Z_{\text{ref}}$ and $Z_{\text{noisy}}$ is significant, and therefore training $W_{K'}$ and $W_{V'}$ of blended attention is needed to bridge $Z_{\text{ref}}$ and $Z_{\text{noisy}}$.

 \begin{figure}[t]
    \centering
    \includegraphics[width=\textwidth]{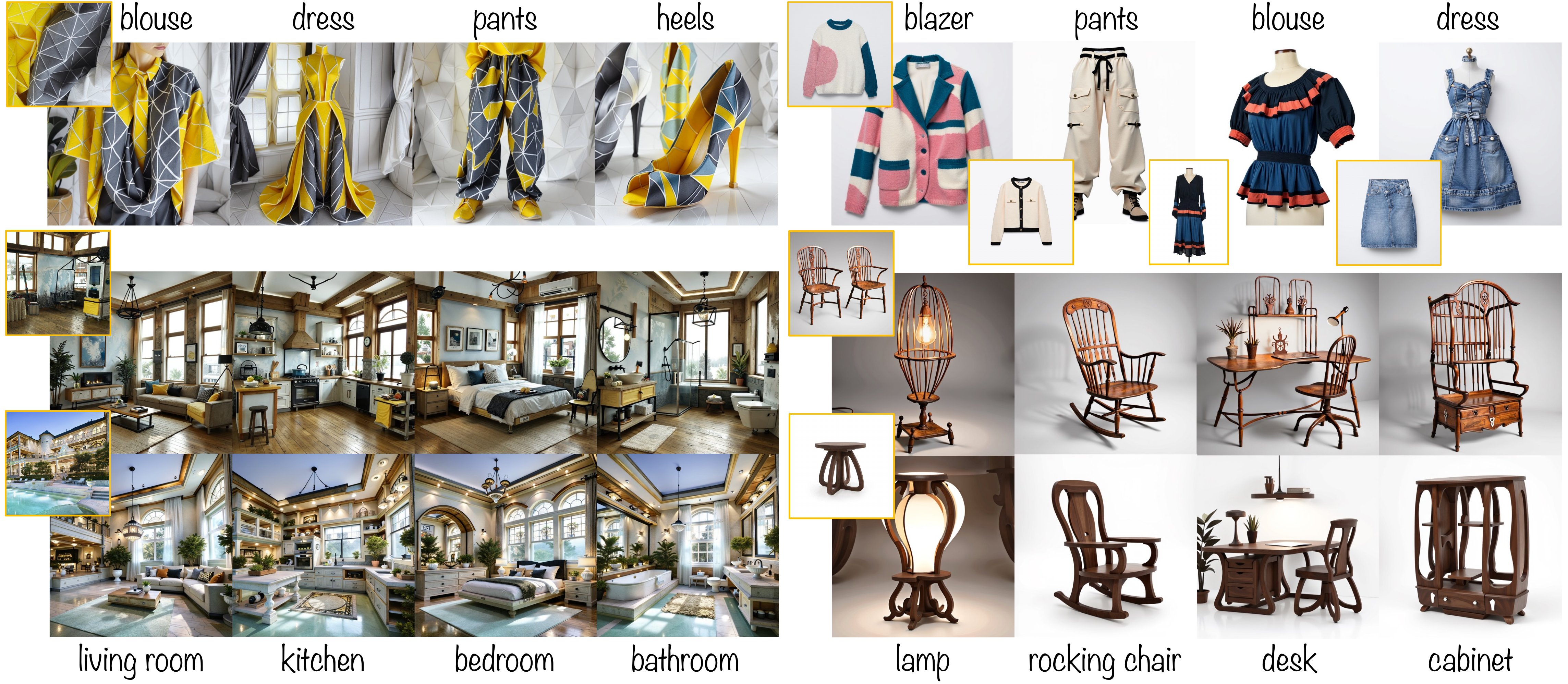}  
    \caption{Feasible design examples when the given visual and textual concepts are semantically close.}
  \label{fig:within category interior and furniture}
  \vspace{-0.2in}
\end{figure}

 \begin{figure}[b]
    \vspace{-0.1in}
    \centering
    \includegraphics[width=\textwidth]{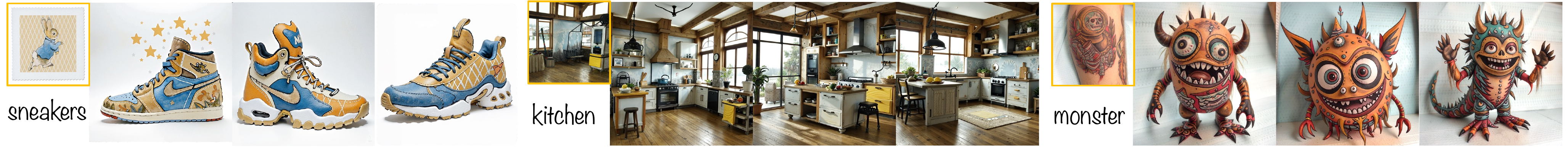}  
    \caption{The results are generated with varying noise.}
  \label{fig:varying noise}
  \vspace{-0.1in}
\end{figure}

\textbf{Fesible design.} 
As shown in the owls with diverse desserts in Fig.~\ref{fig:main 1page figure}, IT-Blender can create experimental design in a realistic way, which can inspire humans. Interestingly, we also observe that IT-Blender can generate feasible design outcomes as well, especially when a reference image is semantically close to the object given in the text prompt. For example, as shown in Fig.~\ref{fig:within category interior and furniture}, given an indoor or outdoor reference image, IT-Blender can generate the target room with surprisingly coherent visual concepts with the given reference image. Furniture or apparel could be another example.

\textbf{Additional results.} 
Given a fixed visual and textual concepts, IT-Blender can generate diverse images with varying random noise, as shown in Fig.~\ref{fig:varying noise}. The creative object generated by IT-Blender can be synthesized in novel scenes with subject-driven models, as shown in Fig.~\ref{fig:mask results and uno} (b).

  


\section{Conclusion}
In this paper, we propose IT-Blender that can augment human creativity by automating the cross-modal conceptual blending process of a real image and text.
First, IT-Blender uses native denoising networks to encode a real reference image to minimize the loss of visual details, with fast inference time. Second, the encoded visual feature is fed into our novel blended attention modules, which are trained to bridge the distribution shift between the clean reference image and the noised generated image. Third, our blended attention modules are built upon the self-attention module, which can disentangle the textual concept and the visual concept by design. In both SD and FLUX, the experiment results demonstrate that IT-Blender outperforms the baselines in blending cross-modal concepts in terms of disentangling cross-modal concepts and preserving textual and visual concepts.
The blending score further verifies the superior performance of IT-Blender in cross-modal conceptual blending. Further discussion of future directions, limitations, and societal impact is provided in Section~\ref{appendix sec: discussion}. We hope that our research will be able to draw attention to the potential of image-generative models to augment human creativity.

\label{sec:conclusion}

{\small
\bibliographystyle{unsrtnat}
\bibliography{reference}
}

\clearpage
\appendix

\section*{Appendices}

\section*{Contents of Appendices}

\begin{figure}[h!]
  \centering
  \includegraphics[width=\textwidth]{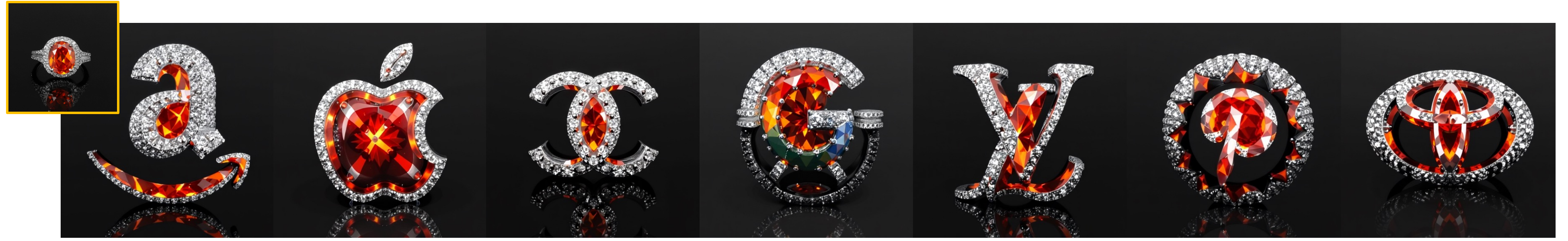}  
  \caption{Stylized brand logos by IT-Blender with FLUX.}
  \label{appendix fig:logo}
\end{figure}

\begin{itemize}
  \item Appendix A: Experiment Settings \dotfill \pageref{appendix sec:experiment settings}
  \item Appendix B: Implementation Details 
  \item Appendix C: Multiple Visual Concepts \dotfill \pageref{appendix sec: multiple visual concepts}

  
  \item Appendix D: Additional Baseline Comparisons 
  \begin{itemize}
      \item D.1: Comparison of Blending Score by ChatGPT (SD and FLUX) \dotfill \pageref{appendix subsec: blending score by chatgpt}
      \item D.2: Qualitative Observation Report (SD) \dotfill \pageref{appendix subsec: limitation of training-free inversion-based method}
      \newline  (A limitation of training-free inversion-based method)
  \end{itemize}
  \item Appendix E: Additional Results and Analysis
  \begin{itemize}
      \item E.1: Effect of $\alpha$ of Blended Attention \dotfill \pageref{appendix subsec:effect of alpha of blended attention}
      \item E.2: Softmax Temperature Control (heuristic for multiple reference images)
      \item E.3: Additional Results \dotfill \pageref{appendix subsec: additional results}

  \end{itemize}
  \item Appendix F: Discussion \dotfill \pageref{appendix sec: discussion}
  \begin{itemize}
      \item F.1: Interesting Future Directions 
      \item F.2: Limitations 
      \item F.3: Societal Impact 
  \end{itemize}
\end{itemize}


\newpage


\section{Experiment Settings}
\label{appendix sec:experiment settings}
\textbf{SD setting.} To evaluate the performance with the baselines in SD, we sample 200 samples per prompt. The 30 prompts that we used are as follows:  

    \texttt{car, bus, bicycle, chair, truck, tank, lamp, 
    handbag, backpack, 
    heels, train, rabbit cartoon character, owl cartoon character, mouse cartoon character, castle,
    headphone, motorcycle, kettle, 
    vacuum, toy airplane, 
    robot, sneakers,
    dragon cartoon character, reindeer cartoon character, alien cartoon character, living room, bathroom, bedroom, kitchen, house}

\textbf{FLUX setting.} We sample 200 samples per prompt. The 20 prompts that we used are as follows:

    \texttt{car, bicycle, chair, lamp, headphone, truck, 
    sneakers, handbag, backpack, t-shir",
    lizard, fish, owl cartoon character, 
    monster cartoon character, dragon,
    living room, kitchen,
    castle,
    3D apple logo, 3D toyota logo}

\section{Implementation Details}
\label{appendix sec:implementation details}

To train IT-Blender with SD 1.5, we use 1 NVIDIA RTX 6000 with a batch size of 16. To train IT-Blender with FLUX, we use 4 NVIDIA L40S GPUs with a total batch size of 16. IT-Blender training and sampling require two streams, as shown in Fig.~\ref{fig:it_blender_overview and imCA}. We simply concatenate them in the batch dimension so that the key-value injections from the reference stream can be easily achieved in each Blended Attention processor. 

We train IT-Blender for 5 epochs with a learning rate of \texttt{1e-5} in SD 1.5. We train IT-Blender for 1-2 epochs with a learning rate of \texttt{2e-5} in FLUX. AdamW~\citep{loshchilov2017decoupled} is used in both settings with \texttt{betas = [0.9, 0.99]} and \texttt{weight\_decay = 0.01}.


\newpage
\section{Multiple Visual Concepts}
\label{appendix sec: multiple visual concepts}
IT-Blender can apply multiple visual concepts from multiple reference images.

The naive way is to add additional imCA terms in Eq.~\ref{eq:blended attention}, e.g., 

\begin{align}
    \text{BA}=\text{SA}(Z_{\text{noisy}}^{(\ell)}) + \alpha \, \text{imCA}&(Z_{\text{noisy}}^{(\ell)},\text{SA}(Z_{\text{ref}_1}^{(\ell)});\frozenweights{W_Q},\trainedweights{W_{K'}},\trainedweights{W_{V'}}) \\ \nonumber
    + \alpha \, &\text{imCA}(Z_{\text{noisy}}^{(\ell)},\text{SA}(Z_{\text{ref}_2}^{(\ell)});\frozenweights{W_Q},\trainedweights{W_{K'}},\trainedweights{W_{V'}}),
    \label{eq:naive multiple visual concepts}
\end{align}
where $Z_{\text{ref}_1}$ and $Z_{\text{ref}_2}$ mean two reference images. However, we empirically observe that the results naively mingles the visual features for each query coordinate, which makes the generated image less conspicuous where the visual feature comes from.

To tackle this problem, we came up with a simple idea; concatenating the multiple reference images in sequence dimension before applying softmax of Attention, e.g., $Q \in \mathbb{R}^{HW \times D}$ and $\{K,V\} \in \mathbb{R}^{2HW \times D}$, when two reference images are used. In this way, BA module can exclusively (not strictly though as it is softmax, not hardmax) fetch the visual features from the multiple reference images. We used this approach to blend multiple visual concepts. More examples are provided below in Fig.~\ref{appendix fig: multiple concepts}.

Another possible way would be to concatenate multiple reference images in height or weight dimension of the reference image, similar to \citet{huang2024context}. 

\begin{figure}[h!]
  \centering
  \includegraphics[width=\textwidth]{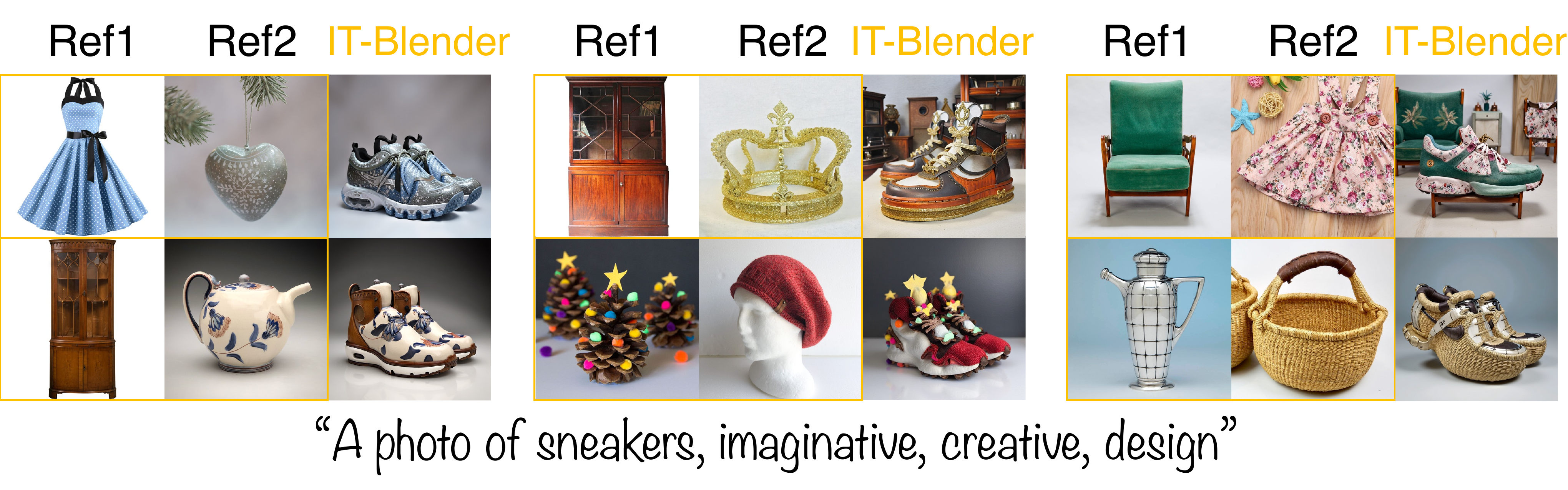}  
  \caption{Examples by IT-Blender with FLUX, generated with multiple reference images.}
  \label{appendix fig: multiple concepts}
\end{figure}

\newpage

\section{Additional Baseline Comparisons}
\label{appendix sec: additional baseline comparisons}

\subsection{Comparison of Blending Score by ChatGPT (SD and FLUX)}
\label{appendix subsec: blending score by chatgpt}
To further measure the blending performance, we use ChatGPT 4.1~\citep{openai2023gpt4} with a detailed rubric,
inspired by the high correlation between human and state-of-the-art LLMs in measuring text and image alignment~\citep{park2024rare}. 

To evaluate, the same samples with the main experiments are used, i.e., the 6000 samples in SD and 4000 samples in FLUX.
The results are as shown below:
\begin{figure}[H]
    \centering
    \begin{subfigure}[b]{0.48\textwidth}
    \includegraphics[width=\linewidth]{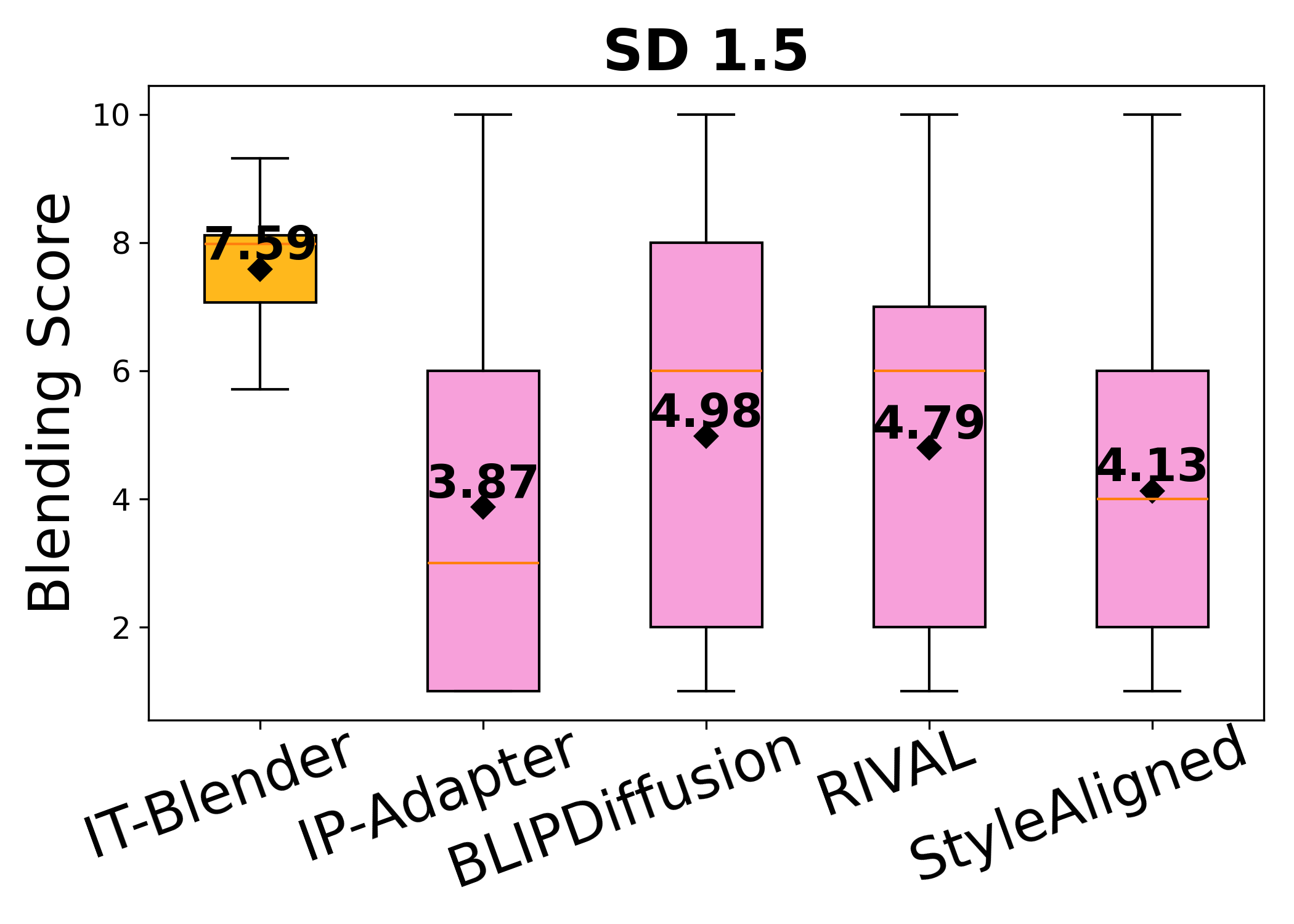}
    \end{subfigure}
    \hfill
    \begin{subfigure}[b]{0.48\textwidth}
    \includegraphics[width=\linewidth]{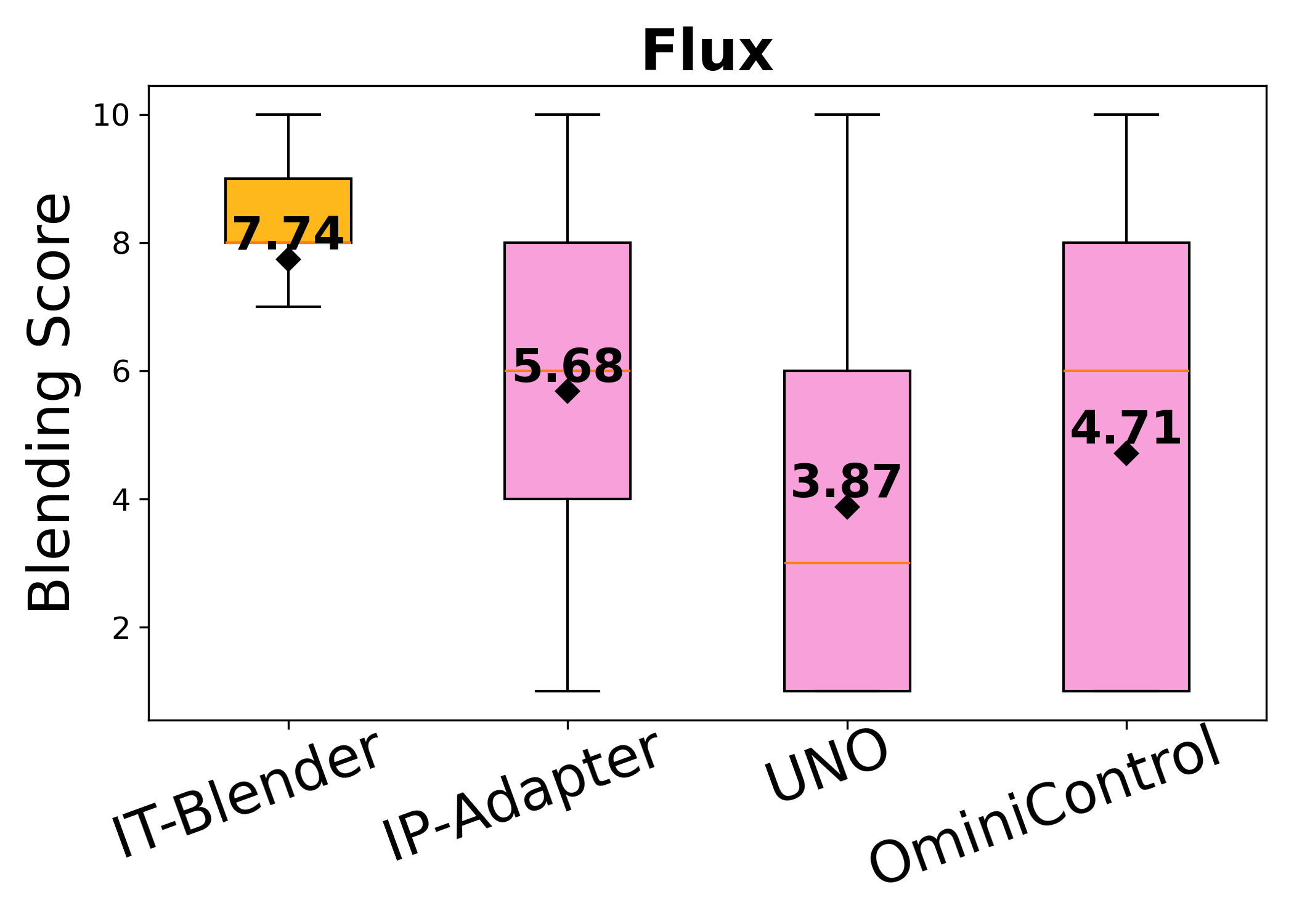}
    \end{subfigure}
    \caption{Visualizations of the blending score comparisons with the baselines in SD (left) and  FLUX (right).}
    \label{fig:blending score}
\end{figure}


Fig.~\ref{fig:blending score} shows the blending score measure by ChatGPT, given a specific rubric.  
As shown in the SD-based and Flux-based results, IT-Blender shows the rigid and best performance with the highest mean and lowest variance. 

According to the rubric, the highest mean around 8 indicates that our blending results have most elements from both inputs, and they are well integrated.

The low variance of IT-Blender indicates that both concepts are consistently blended in a plausible way, without failed or unbalanced integration.

We further visualize the top 10\%, 50\% (median), and 90\% samples in terms of the blending score in Fig.~\ref{appendix fig: blending score visualization, top5 percent} and Fig.~\ref{appendix fig: blending score visualization, bottom5 percent}. The high blending scores around 8-9 show decent performance in blending visual and textual concepts while the low blending scores around 1-2 show poor performance, e.g., only applying one concept or blending cross-modal concepts weakly.

\begin{figure}[h!]
  \centering
  \includegraphics[width=\textwidth]{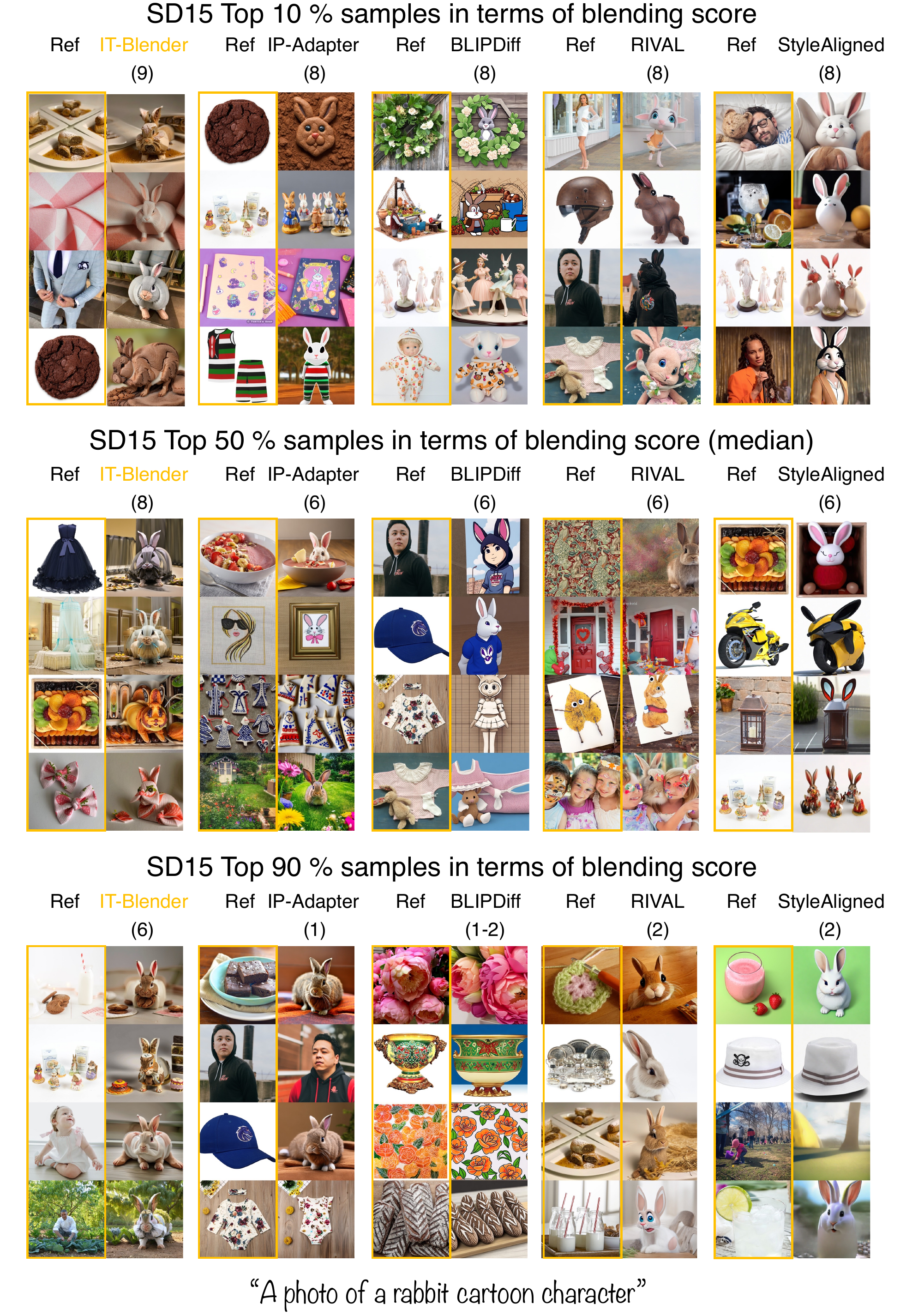}  
  \caption{Visualization of top 10\%, 50\%, and 90\% samples in terms of blending score (SD). The numbers below each baseline name indicate the blending scores the displayed samples got.}
  \label{appendix fig: blending score visualization, top5 percent}
\end{figure}

\begin{figure}[h!]
  \centering
  \includegraphics[width=\textwidth]{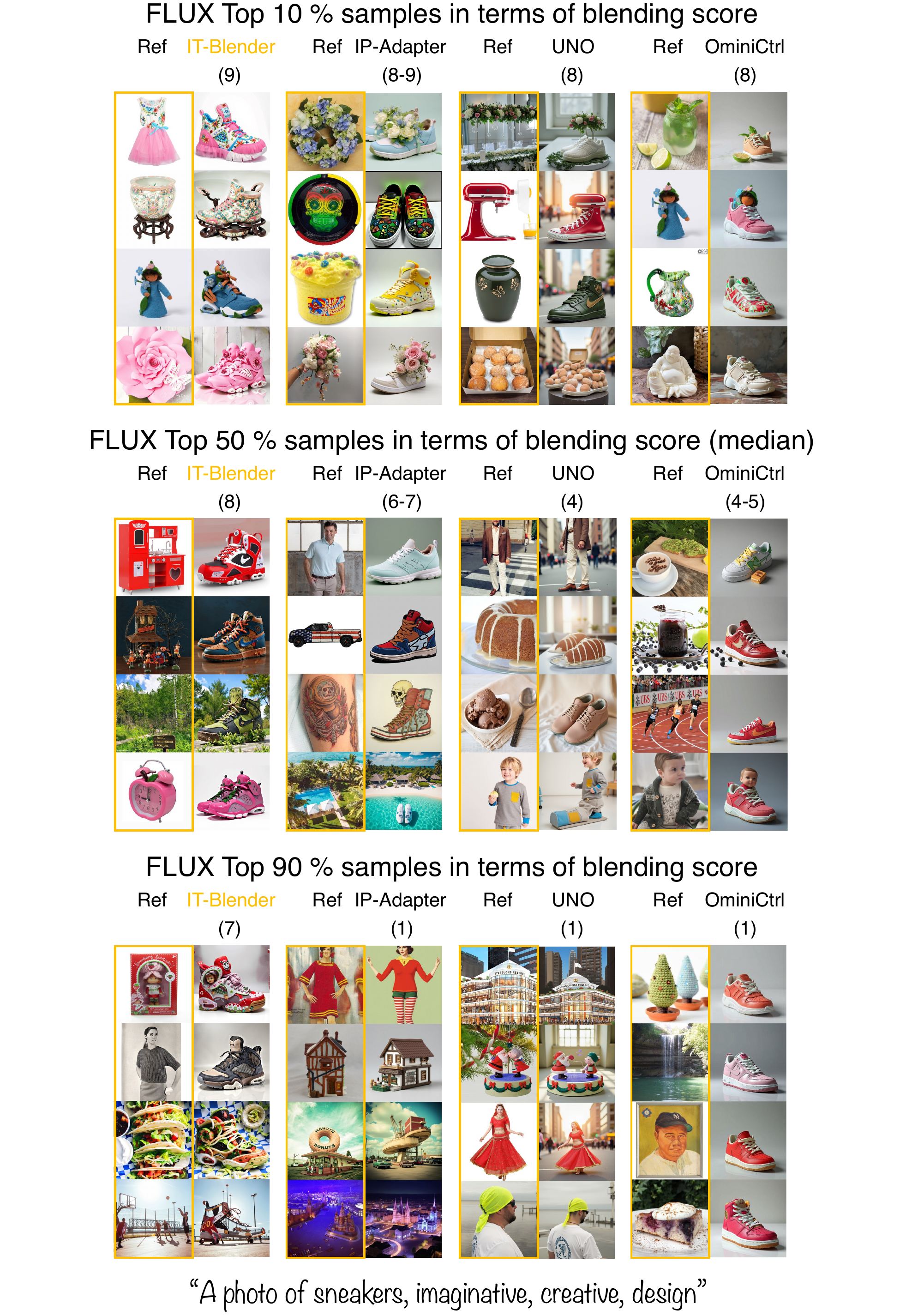}  
  \caption{Visualization of top 10\%, 50\%, and 90\% samples in terms of blending score (FLUX). The numbers below each baseline name indicate the blending scores the displayed samples got.}
  \label{appendix fig: blending score visualization, bottom5 percent}
\end{figure}

\clearpage

\paragraph{Query for measuring blending score.}
The prompt we used to measure the blending score is as follows:

\texttt{You are a helpful assistant who evaluates how well textual and visual concepts are blended in the image generation process.
            The object in the given first image is conceptually blended result given the text prompt and the second reference image. 
            Text determines semantics while the reference image determines visual concepts such as texture, material, color, and local shape.
            Evaluate how closely the visual concept in the provided image aligns with the textual concept in the text prompt and the visual concept from the second image. 
            Identify significant overlaps or discrepancies in terms of global object shape, local shape, appearance, texture, material, color, and all the detailed visual components. 
            Analyze the conceptual similarity between the first provided generated image and the text prompt: [PROMPT]. 
            You also need to consider the conceptual similarity between the first provided generated image and the second provided reference image.
            Provide a concise explanation for your evaluation. 
            Note that we are evaluating cross modal conceptual blending, and thus if one of the crossmodal concepts does not present in the generated image, it has to be considered as failed, even though the first image perfectly matches the second image.}

\texttt{First image: [GEN\_IMAGE]}

\texttt{Second image: [REF\_IMAGE]}

\texttt{The object in the given first image is conceptually blended result given the text prompt and the second image. Evaluate how closely the visual concept in the provided image aligns with the textual concept in the text prompt and the visual concept from the second image. Identify significant overlaps or discrepancies in terms of shape, appearance, composition, and overall impression.
Provide a concise explanation for your evaluation.}

\texttt{Give a score from 1 to 10, according to the following criteria:}

\texttt{10	Perfect conceptual integration: The generated image seamlessly incorporates all core semantic and stylistic elements from both the text and the visual concept. There’s no ambiguity in the fusion; it reflects a deep, coherent synthesis of the two modalities.}

\texttt{9	Near-perfect integration: Strong conceptual blending with only extremely minor details or subtleties missing from either modality. The result is still fully coherent and creatively unified.}

\texttt{8	Excellent with minor trade-offs: Most elements from both inputs are present and well-integrated, but one or two key aspects may be simplified. The conceptual overlap is still meaningful.}

\texttt{7	Very good blend, slightly unbalanced: Clear depiction of both concepts with small discrepancies—e.g., one modality slightly dominates the fusion. Still communicates a unified concept.}

\texttt{6	Mostly present, but noticeable gaps: Both modalities are represented, but some important attributes (e.g. color, pose, key terms, or symbolic features) are missing or only vaguely suggested.}

\texttt{5	Moderate representation: Some elements from both text and image are depicted, but several key parts are ignored or distorted. The blend may feel partial or underdeveloped.}

\texttt{4	Unbalanced or sparse blend: One modality is clearly underrepresented or the blend feels superficial. Visuals may include token features from one source without meaningful synthesis.}

\texttt{3	Weak conceptual integration: Few recognizable aspects from both text and image appear; blending feels incomplete or accidental rather than intentional.}

\texttt{2	Minimal blending: Image mostly reflects one modality, with only token or confused reference to the other. Viewers may struggle to infer any deliberate fusion.}

\texttt{1	Failed integration: Generated image does not meaningfully reflect either the textual concept or the visual input. No clear blending is achieved.}

\texttt{Provide your score and explanation (within 20 words) in the following format:
    \#\#\# SCORE: score
    \#\#\# EXPLANATION: explanation
}

\paragraph{Explanations provided by ChatGPT for blending score.}
The explanations for the lowest row for each of top 10\%, 50\%, and 90\% are as follows: 

\begin{itemize}
    \item SD15
    \begin{enumerate}    
        \item IT-Blender
        \begin{enumerate}
            \item (Top 10\%) ``The rabbit cartoon's form is clear and well-blended with the cookie's texture and color, though small stylization remains.''
            \item (Top 50\%) ``The image blends a rabbit cartoon character with fabric and color from the bows, but lacks full cartoon stylization.''
            \item (Top 90\%) ``The rabbit shows garden background and chef attire (from the reference), but lacks strong cartoon character cues from text.''
        \end{enumerate}
        \item IP-Adapter
        \begin{enumerate}
            \item (Top 10\%) ``Rabbit cartoon is fully integrated; clothing shows reference outfit’s colors and stripes, but lacks emblem and exact shape.''
            \item (Top 50\%) ``The rabbit matches the text but is realistically rendered, not cartoon-like; the garden environment strongly reflects the reference image.''
            \item (Top 90\%) ``The generated image contains only the visual reference style, with no trace of the "rabbit cartoon character" concept.''
        \end{enumerate}
        \item BLIP-Diffusion
        \begin{enumerate}
            \item (Top 10\%) ``''
            \item (Top 50\%) ``The generated image uses the reference's color and texture but misses key elements of "rabbit cartoon character" in pose and style.''
            \item (Top 90\%) ``The generated image reflects only the bread visual, lacking any "rabbit cartoon character" elements from the text prompt.''
        \end{enumerate}
        \item RIVAL
        \begin{enumerate}
            \item (Top 10\%) ``Strong rabbit cartoon character blending, large expressive eyes and pose, but clothing details lack reference's color and pattern accuracy.''
            \item (Top 50\%) ``The generated image shows a real rabbit and kids with cartoonish face paint, but lacks full cartoon character integration.''
            \item (Top 90\%) ``Image matches the rabbit cartoon text prompt, but shows almost no visual or conceptual blending with the milk bottles.''
        \end{enumerate}
        \item StyleAligned
        \begin{enumerate}
            \item (Top 10\%) ``Strong blend; rabbit cartoon matches text while outfit, pose, and background clearly reflect the reference image. Minor stylization only.''
            \item (Top 50\%) ``Rabbit forms and some cartoon stylization are present, but features are indistinct and visuals are muddled.''
            \item (Top 90\%) ``''
        \end{enumerate}
    \end{enumerate}
    \item FLUX
    \begin{enumerate}
    \item IT-Blender
    \begin{enumerate}
        \item (Top 10\%) ``The sneaker perfectly incorporates the reference flowers’ color, material, and shape, with only minor textural differences from the reference.''
        \item (Top 50\%) ``Sneakers integrate clock's pink color, shiny texture, and green leaf elements, but clock face and apple shape are abstracted.''
        \item (Top 90\%) ``Strong sneaker form integrates basketball court elements and colors, but local sneaker details and textures are somewhat abstracted.''
    \end{enumerate}
    \item IP-Adapter
    \begin{enumerate}
        \item (Top 10\%) ``The sneaker integrates floral elements—shape and details—from the bouquet while retaining clear sneaker form, with only minor detail loss.''
        \item (Top 50\%) ``Sneakers (text) are clearly integrated into the villa pool scene (reference), but sneakers' material/style don't borrow villa textures.''
        \item (Top 90\%) ``No sneakers are present; the image depicts buildings and cityscape, failing both text and visual blending criteria.''
    \end{enumerate}
    \item UNO
    \begin{enumerate}
        \item (Top 10\%) ``Sneaker shape is clear and main structure matches "sneakers", but donut texture dominates, slightly stylizing the footwear concept.''
        \item (Top 50\%) ``Only the clothing from the reference is blended; no real sneaker shape from the text is present, making fusion superficial.''
        \item (Top 90\%) ``The generated image only depicts a man with a yellow headscarf, not sneakers; it fails cross-modal blending.''
    \end{enumerate}
    \item OminiControl
    \begin{enumerate}
        \item (Top 10\%) ``The sneaker adopts the Buddha statue's ivory color, material, and some smooth texture, but lacks significant Buddha-specific shapes.''
        \item (Top 50\%) ``The sneaker incorporates a doll's head, referencing the baby, but lacks deeper integration of baby features, mainly merging objects.''
        \item (Top 90\%) ``The generated image is a sneaker, matching only the text prompt, with no visual or conceptual blending of the cake reference.''
    \end{enumerate}
\end{enumerate}

\end{itemize}

\clearpage

\begin{figure}[h!]
  \centering
  \includegraphics[width=\textwidth]{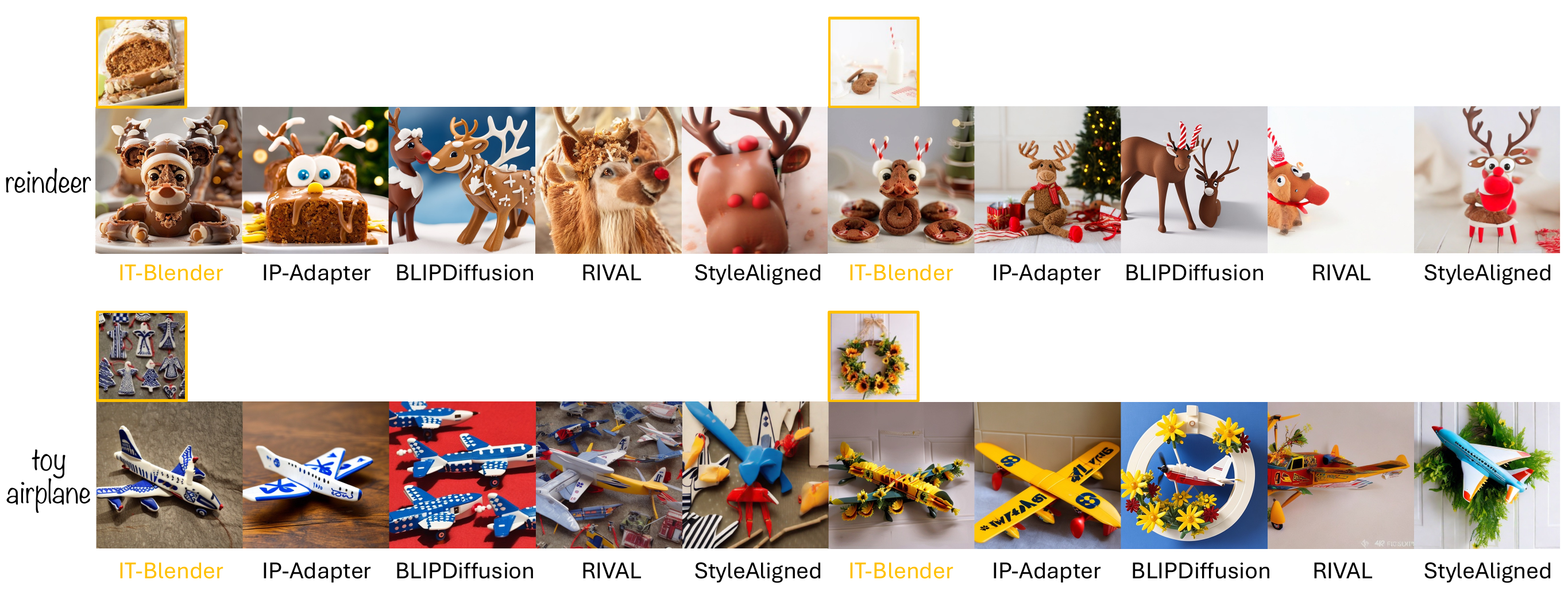}  
  \caption{Additional qualitative comparisons (SD).}
  \label{appendix fig: additional SD qualitative comparison with baselines}
\end{figure}

\subsection{Qualitative Observation Report (SD)}
\label{appendix subsec: limitation of training-free inversion-based method}

we observe that the training-free inversion-based baselines sometimes lie off the manifold, so the results are not realistic when cross-modal concepts are blended. We think this is an inherent limitation of training-free methods (in exchange for the benefit of ``training free''), which intervene in the sampling trajectory. As shown in Fig.~\ref{appendix fig: additional SD qualitative comparison with baselines}, The training-free methods RIVAL and StyleAligned sometimes unrealistically blend the results. The encoder-based baselines IP-Adapter and BLIP-Diffusion often miss the text prompt while the generated results are realistic. IT-Blender combines the benefits, consistently and realistically blending both concepts.

\newpage
\section{Additional Results and Analysis}
\label{appendix sec: additional results}
\subsection{Effect of $\alpha$ of Blended Attention}
\label{appendix subsec:effect of alpha of blended attention}
\begin{figure}[h!]
  \centering
  \includegraphics[width=\textwidth]{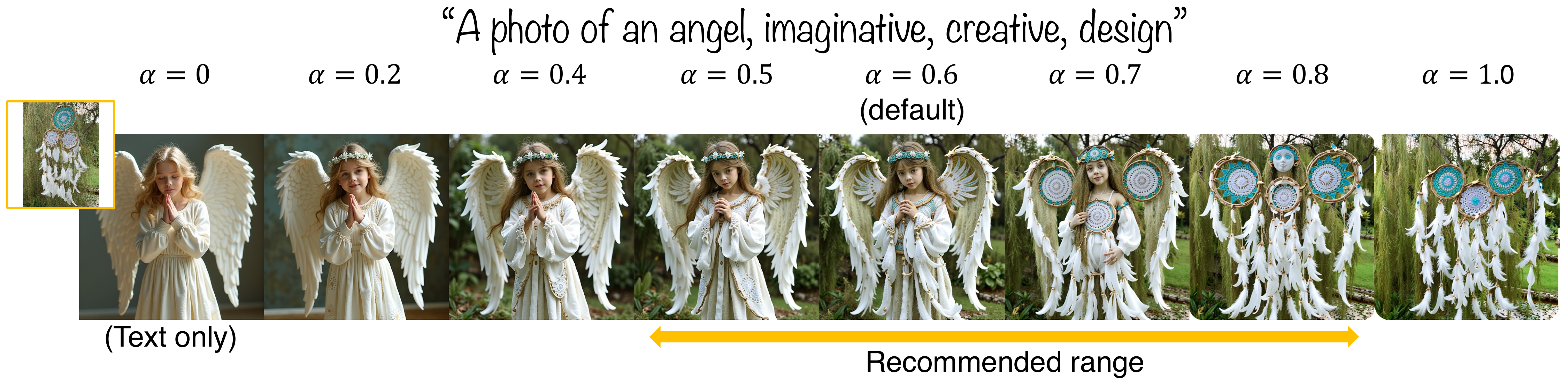}  
  \caption{Visualization of the effect of alpha in blended attention with FLUX.}
  \label{appendix fig:varying_alpha}
  \vspace{-0.2in}
\end{figure}

We visualize the effect of $\alpha$ of Eq.~\ref{eq:blended attention} in Fig.~\ref{appendix fig:varying_alpha}. When $\alpha=0$, no effect is applied as the imCA term in Eq.~\ref{eq:blended attention} becomes zero out. From left to right, as $\alpha$ increases, we can see that the visual concepts are more blended into the generated image. We empirically found that $\alpha=0.6$ is the best way to get the most natural blend results. However, depending on the user's intention, $\alpha \in [0.5,0.8]$ is also good to go with. Especially when reference images and the text prompt are semantically close, $\alpha>0.6$ can be effective, as shown in some of the results in section~\ref{appendix subsec: additional results}.
\label{appendix subsec: varying alpha}

\subsection{Softmax Temperature Control (heuristic for multiple reference images)}
\label{appendix subsec: softmax temperature control}

\begin{figure}[h!]
  \centering
  \includegraphics[width=\textwidth]{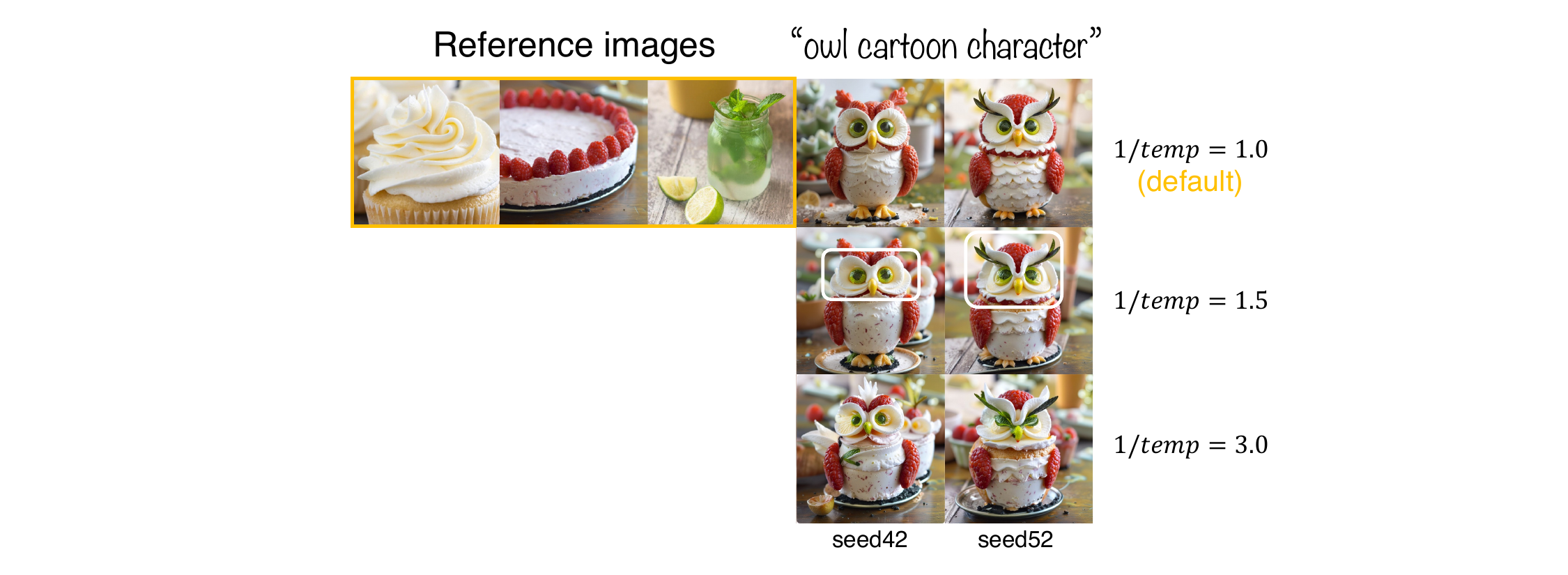}  
  \caption{Visualization of the effect of temperature on the attention mask. Lower temperatures result in less ambiguous and more conspicuous application of visual concepts in exchange for the image fidelity. We empirically observe that $1<1/temp<1.5$ can mitigate the ambiguity when multiple reference images yield ambiguous mixtures of visual concepts.}
  \label{appendix fig:temperature heuristic}
\end{figure}

We empirically observe that applying low temperature to the logits before applying softmax can sharpen the softmax distribution, possibly helping to prevent ambiguous mixtures of visual concepts in exchange for image fidelity. The attention formulation with the temperature can be represented as:

\begin{equation}
    \text{Attention}(Q,K,V;temp) = \text{softmax}\left( \frac{QK^T}{\sqrt{d_k} \cdot temp} \right) V.
\label{eq:attention with temperature}
\end{equation}
$1/temp=1.0$ indicates the default attention mask while $(1/temp)>1.0$ means the attention mask with a sharpened distribution. As shown in the white boxes in Fig.~\ref{appendix fig:temperature heuristic}, applying lower temperature can make the generated results have more conspicuous visual concept. For example, it is vague to determine whether the owl's eyes when $1/temp=1$ come from the first reference image or the second reference image. On the other hand, when $1/temp=1.5$, we can see that the cream texture of the first reference image is drawn more clearly in the generated images. We empirically observe that setting 
$1<1/temp<1.5$ can help mitigate ambiguous mixtures of visual concepts when using multiple reference images. However, note that values of $1/temp>1.0$ may degrade image fidelity.

\clearpage
\subsection{Additional Results}
\label{appendix subsec: additional results}
In this section, we show additional feasible use cases of IT-Blender in diverse design fields. Reference images and a text prompt are semantically close. More additional results with the original resolution can be found on our project page: \trainedweights{\url{https://imagineforme.github.io/}}. 

\begin{figure}[h!]
  \centering
  \includegraphics[width=0.9\textwidth]{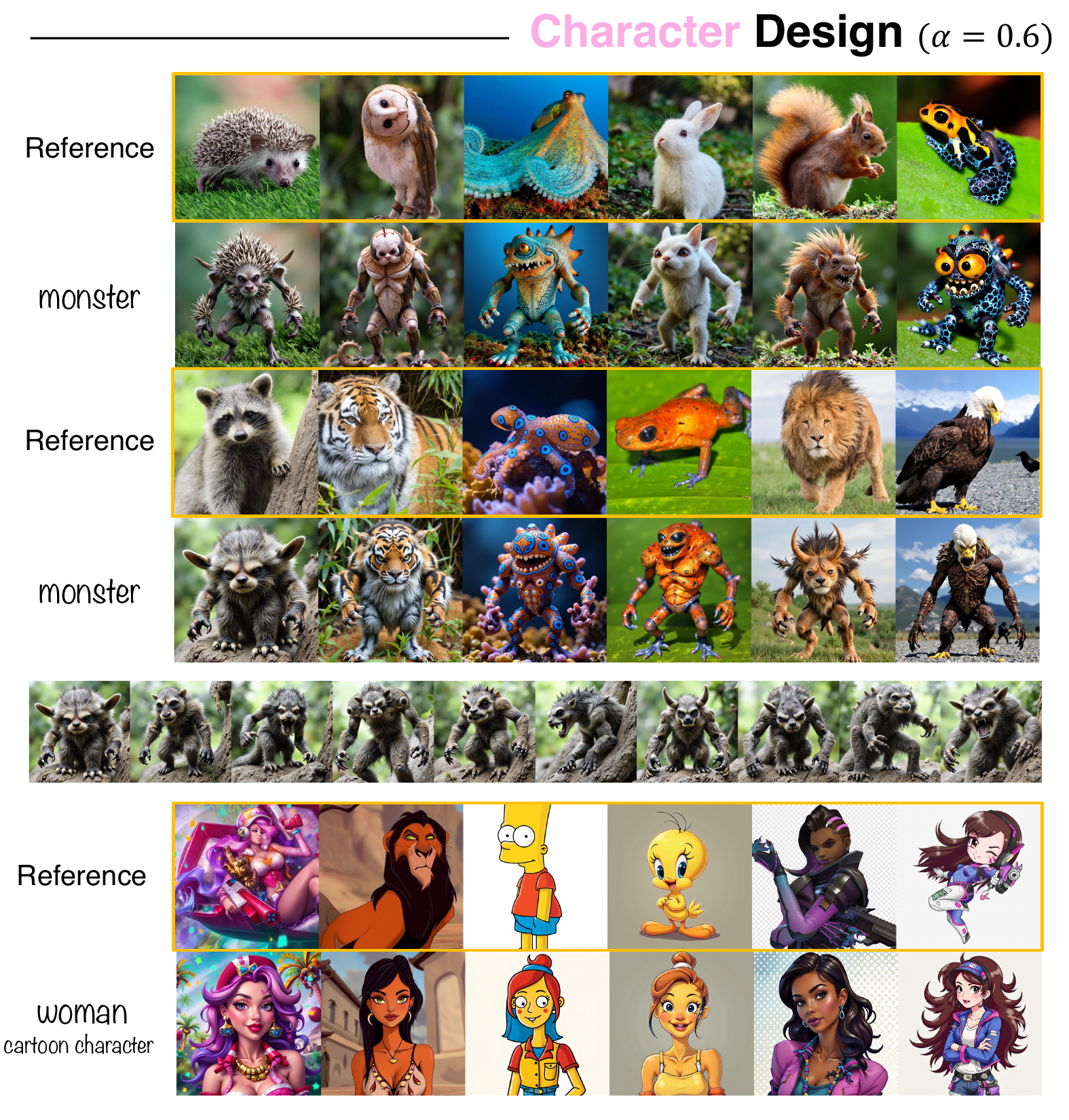}  
  \caption{Feasible character design examples by IT-Blender with FLUX.}
  \label{appendix fig:feasible character design}
\end{figure}

\begin{figure}[h!]
  \centering
  \includegraphics[width=0.9\textwidth]{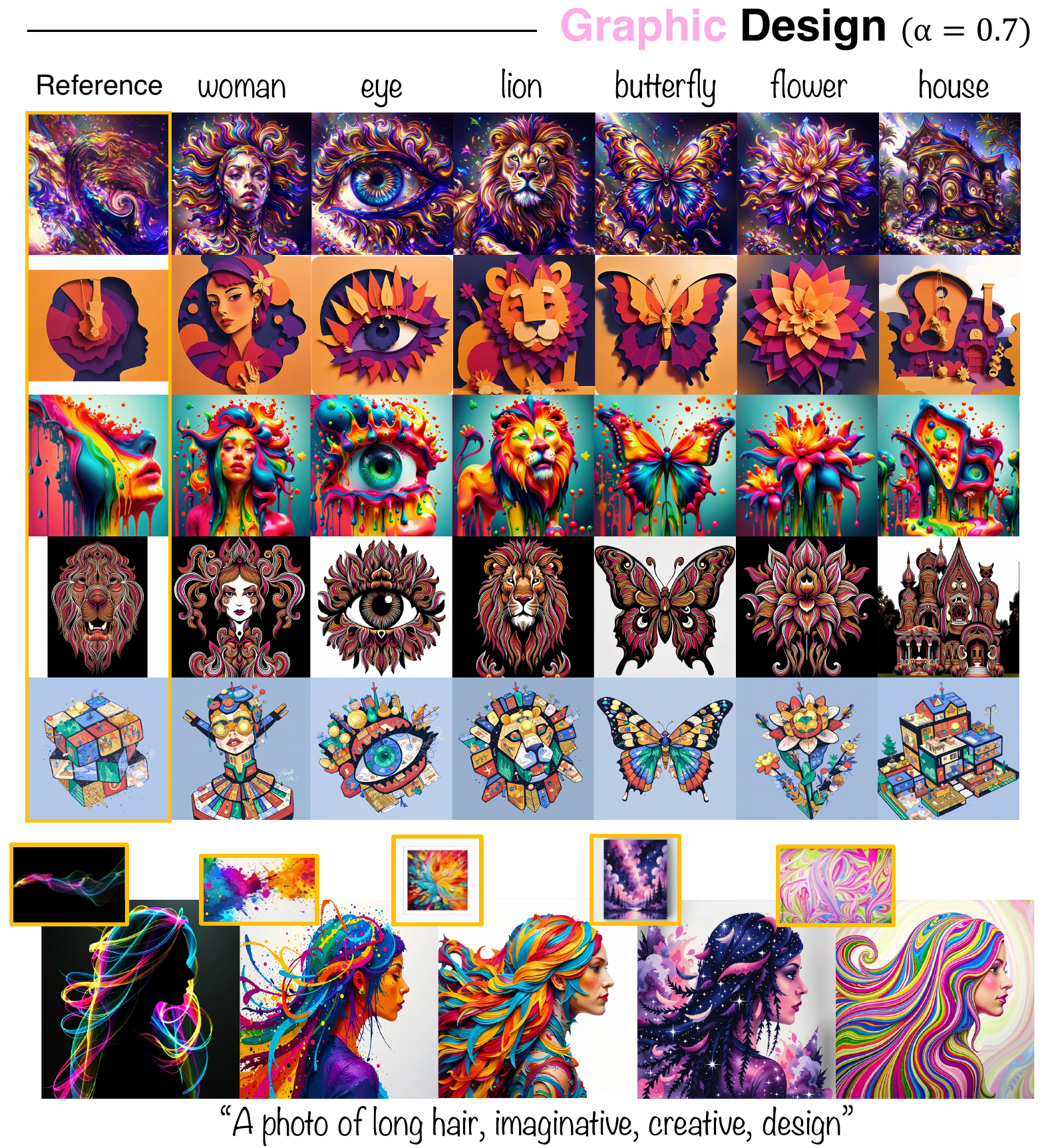}  
  \caption{Feasible graphic design examples by IT-Blender with FLUX.}
  \label{appendix fig:feasible graphic design}
\end{figure}

\begin{figure}[h!]
  \centering
  \includegraphics[width=0.9\textwidth]{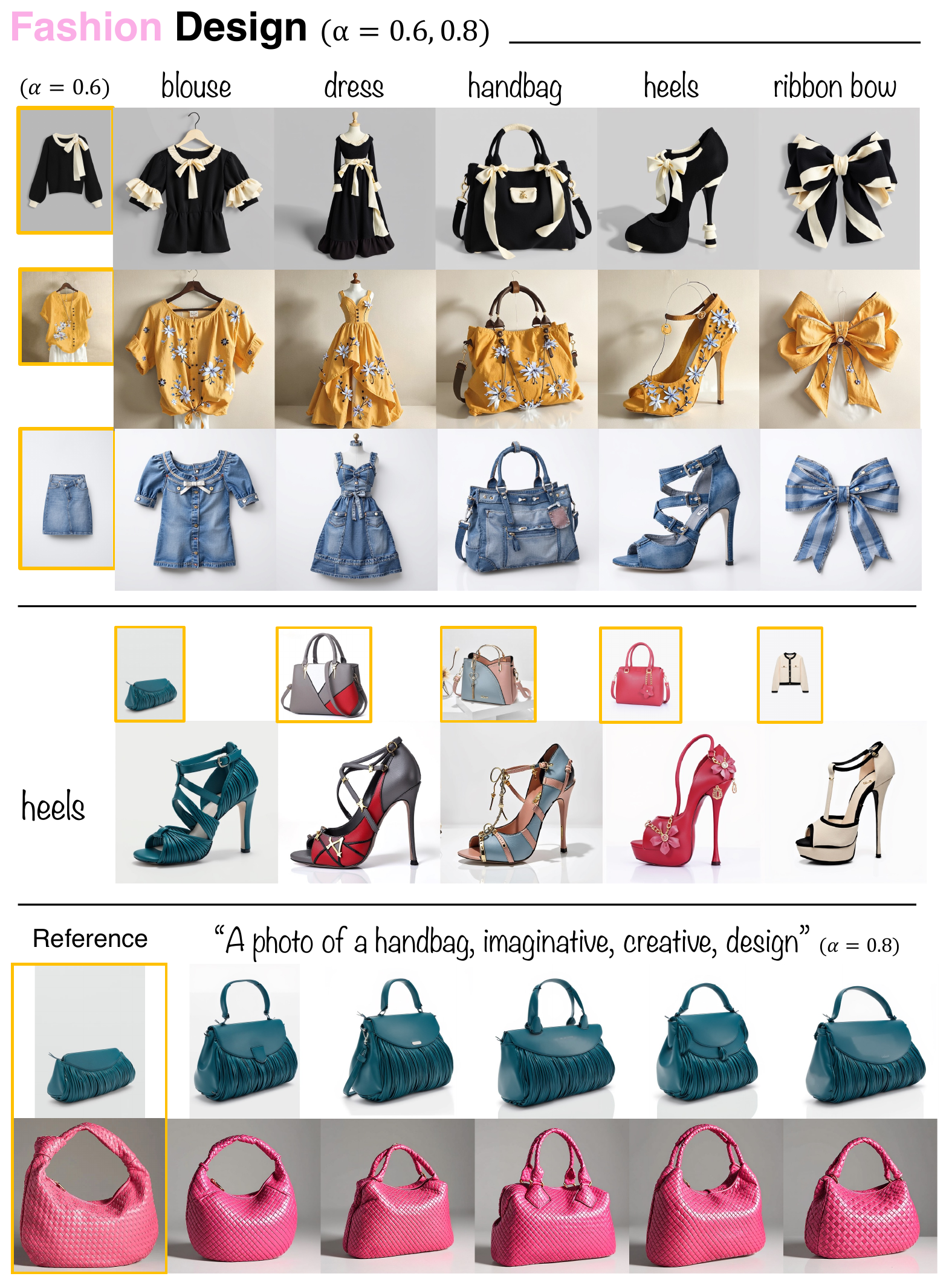}  
  \caption{Feasible fashion design examples by IT-Blender with FLUX.}
  \label{appendix fig:feasible fashion design}
\end{figure}

\begin{figure}[h!]
  \centering
  \includegraphics[width=0.9\textwidth]{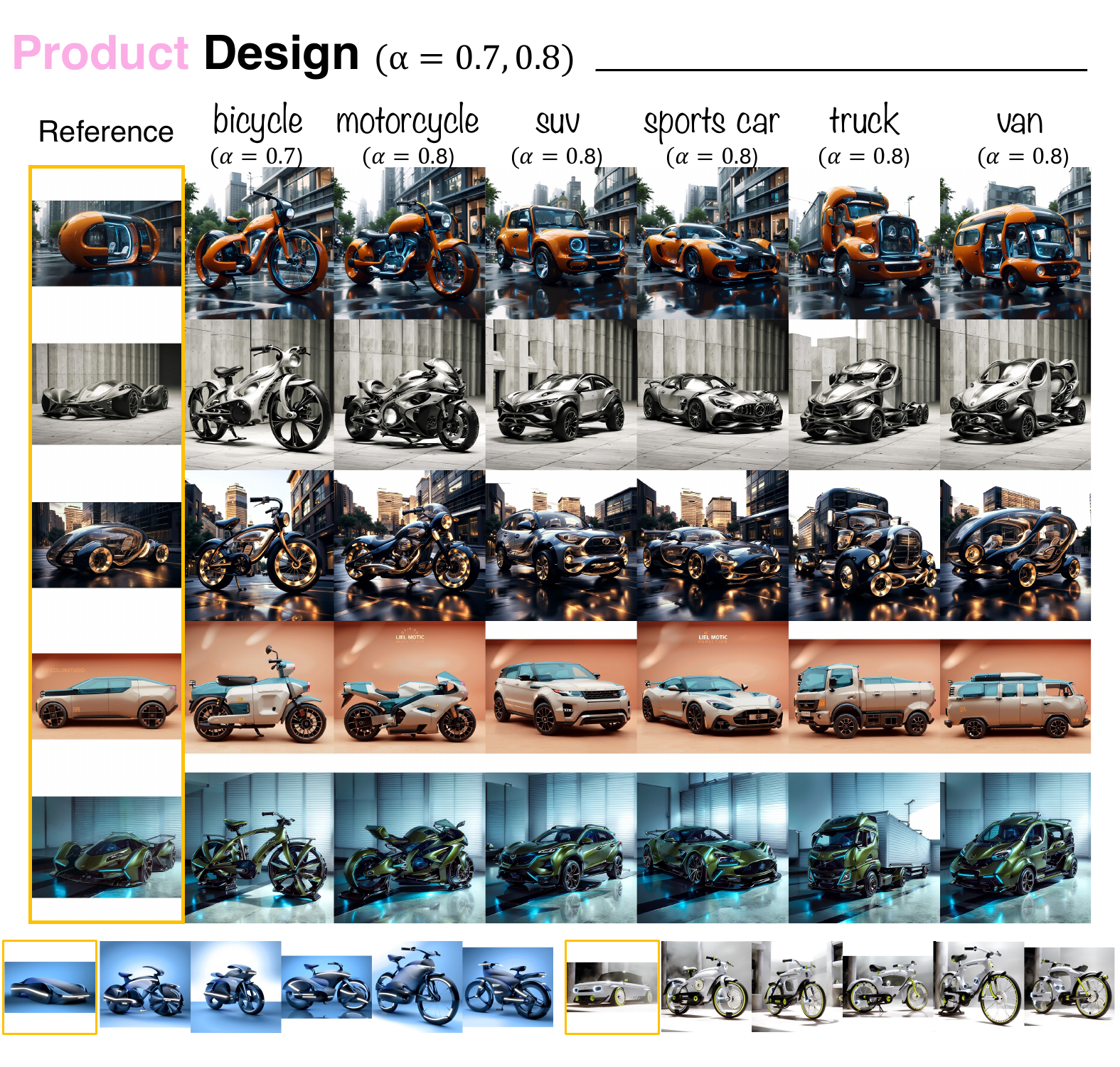}  
  \caption{Feasible product design examples by IT-Blender with FLUX.}
  \label{appendix fig:feasible product design}
\end{figure}

\begin{figure}[h!]
  \centering
  \includegraphics[width=0.9\textwidth]{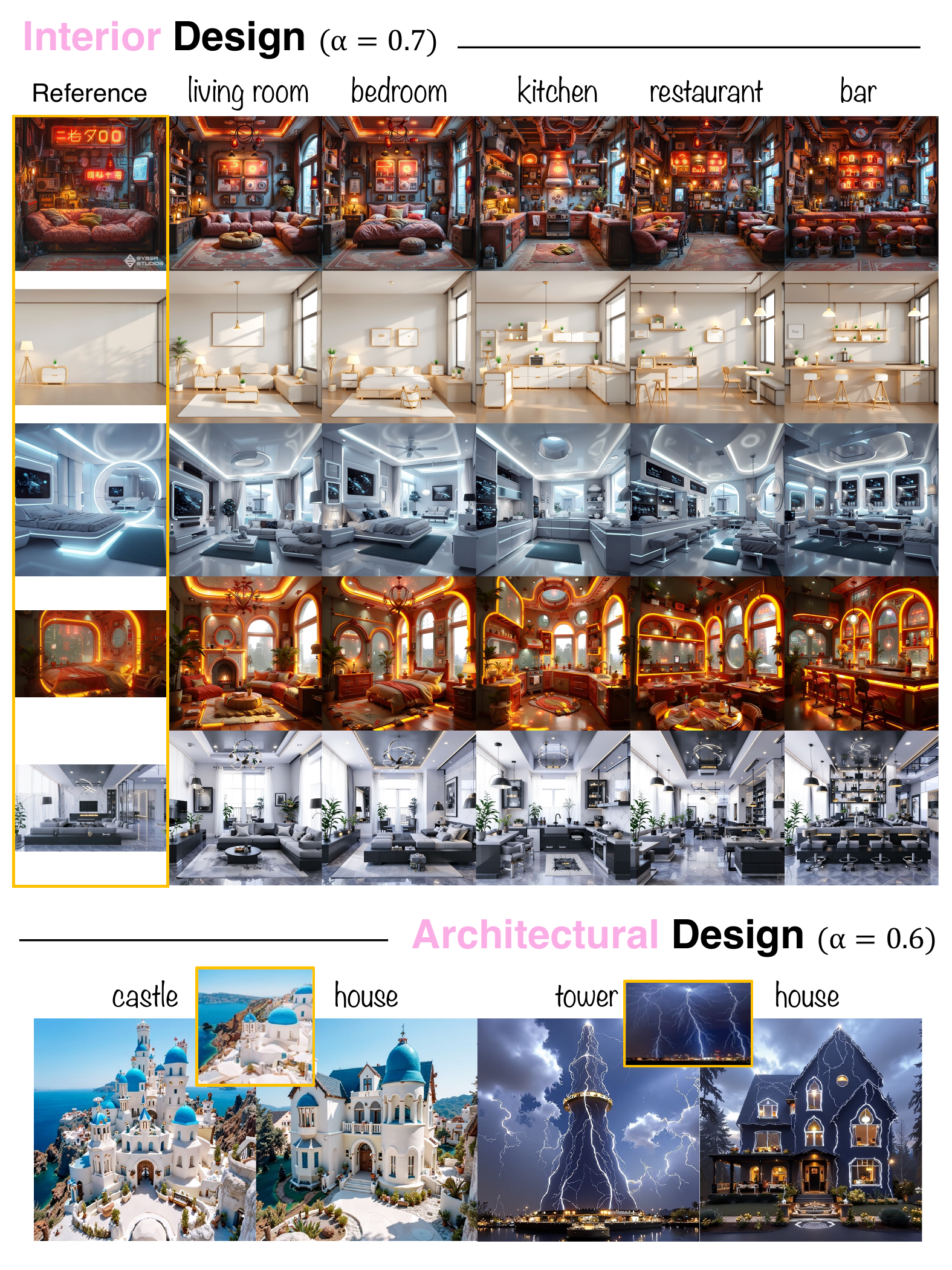}  
  \caption{Feasible interior and architectural design examples by IT-Blender with FLUX.}
  \label{appendix fig:feasible inoutterior design}
\end{figure}

\begin{figure}[h!]
  \centering
  \includegraphics[width=0.9\textwidth]{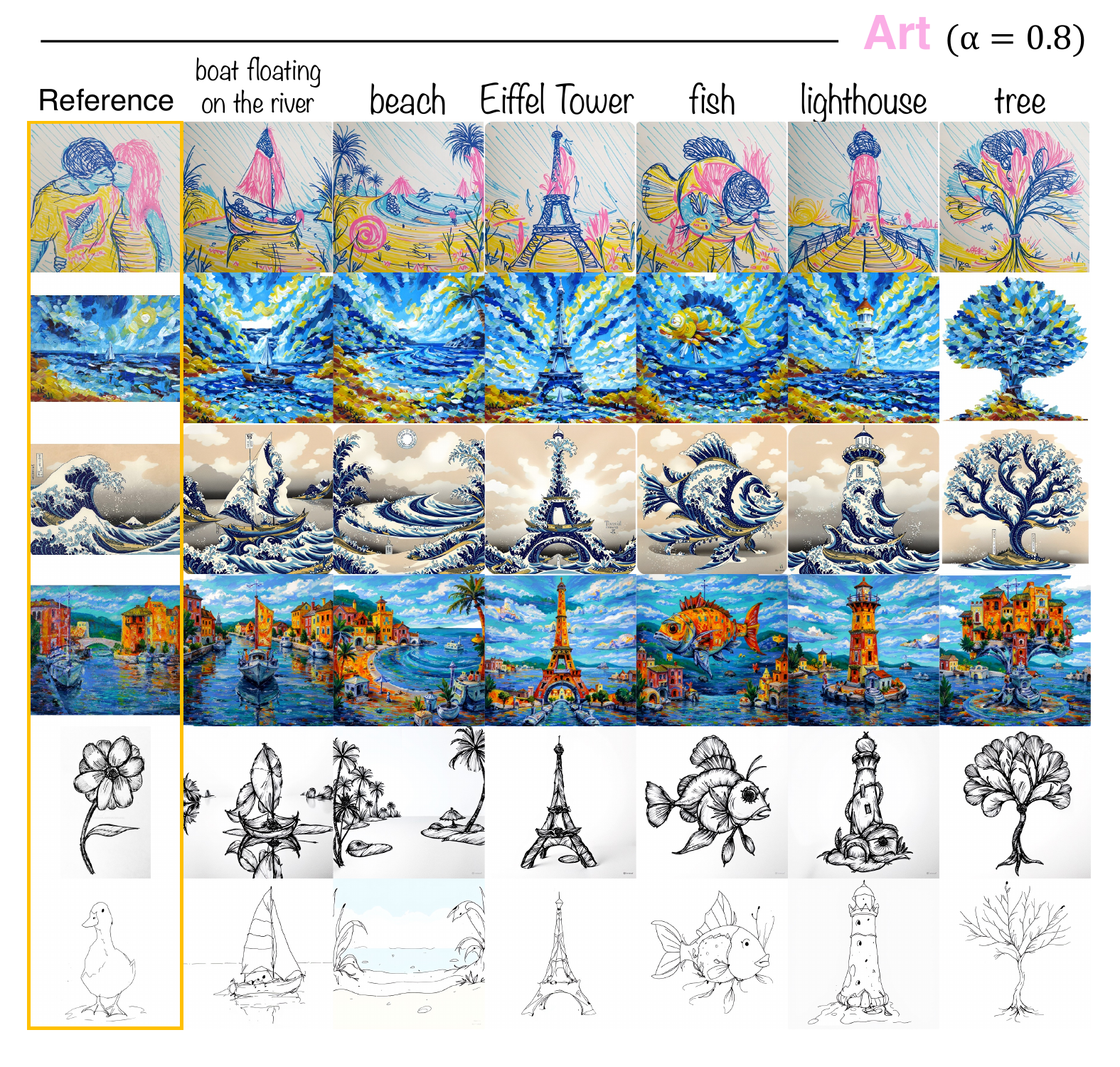}  
  \caption{Feasible art examples by IT-Blender with FLUX.}
  \label{appendix fig:feasible inoutterior design}
\end{figure}

\clearpage

\section{Discussion}
\label{appendix sec: discussion}

\subsection{Interesting Future Directions}
\label{appendix subsec: interesting future directions}
Our proposed blended attention module learns to be specialized in retrieving semantic correspondence between the real image and the generated image, and it combines the visual concept with the text-guided generated image in a plausible way. We believe this technique can be useful in other creativity fields as well, such as music, text and video. For example, suppose we have music generative models. Given an arbitrary table tapping sound, the generated music would have the table tapping sound as a central theme in a plausible way. In another case, suppose that we have text generative models. Given a dialogue from a specific target person as input to the BA module, the generated text will be personalized for that individual.

\subsection{Limitations}
\label{appendix subsec: limitations} Even though IT-Blender shows impressive performance in cross modal conceputal blending, there can be several limitations. First, visual concept subtraction is not working well. It would be interesting if visual concept subtraction could be achieved. 

Second, the global shape variation of the generated objects is limited. In IT-Blender, the semantics of the generated image are determined by a textual condition, and the visual concepts, such as color, texture, local shape, and material, are determined by the reference image. As can be seen in our experiments, the visual concepts can be applied with a large variation. However, the variation of the global shape (i.e., the object) is relatively limited, e.g., given ``heels'', the results literally look like ``heels''. We believe human designers can imagine global shape as well, which we think can be the gap with IT-Blender.

Third, there is room for fully supporting human designers. The aesthetic (i.e., how it looks) is one of the most important features of design, for which IT-Blender can significantly help human designers. However, a good human designer can consider many other features, such as functionality, usability, durability, affordability, and cultural relevance, for which IT-Blender may not be helpful. Further exploration and research are needed for AI that can consider all the important features in design.

\subsection{Societal Impact}
\label{appendix subsec: societal impact}
\textbf{Positive societal impact.} IT-Blender can augment human creativity, especially for people in creative industries, e.g., design and marketing. With IT-Blender, designers might be able to have better final design outcome by exploring wide design space in the ideation stage. 

\textbf{Negative societal impact.} As shown in Fig.~\ref{fig:within category interior and furniture} and Fig.~\ref{appendix fig:feasible character design}-\ref{appendix fig:feasible inoutterior design}, IT-Blender can be used to apply the design of an existing product to the new products. The user must be aware of the fact that they can infringe on the company's intellectual property if a specific texture pattern or material combination is registered. We encourage users to use IT-Blender to augment creativity in the ideation stage, rather than directly having a final design outcome. 


\end{document}